%% file: sample-sigconf.tex
\documentclass[sigconf]{acmart}
\AtBeginDocument{%
  }

\copyrightyear{2025}
\acmYear{2025}
\setcopyright{acmlicensed}\acmConference[KDD '25]{Proceedings of the 31st ACM SIGKDD Conference on Knowledge Discovery and Data Mining V.2}{August 3--7, 2025}{Toronto, ON, Canada}
\acmBooktitle{Proceedings of the 31st ACM SIGKDD Conference on Knowledge Discovery and Data Mining V.2 (KDD '25), August 3--7, 2025, Toronto, ON, Canada}
\acmDOI{10.1145/3711896.3737010}
\acmISBN{979-8-4007-1454-2/2025/08}

\usepackage[utf8]{inputenc} % allow utf-8 input
\usepackage[T1]{fontenc}    % use 8-bit T1 fonts
\usepackage{hyperref}       % hyperlinks
\usepackage{url}            % simple URL typesetting
\usepackage{amsfonts}       % blackboard math symbols
\usepackage{amsmath}
\usepackage{amsthm}
\usepackage{booktabs} 
\usepackage{nicefrac}       % compact symbols for 1/2, etc.
\usepackage{microtype}      % microtypography
\usepackage{xcolor}         % colors
\input{math_commands.tex}
\usepackage{wrapfig}
\usepackage{subfigure}
\usepackage{graphicx}
\usepackage{bbm}
\usepackage{bbold}
\usepackage{listings}
\usepackage{algorithm} 
\usepackage{algpseudocode} 

\usepackage{xspace}
\usepackage{subcaption}
\usepackage{multirow}
\usepackage{multicol}
\usepackage{chemfig}
\setchemfig{
    atom style={scale=0.3},    
    bond style={line width=0.4pt},
    compound style={scale=0.5}
}

\definecolor{codegreen}{rgb}{0,0.6,0}
\definecolor{codegray}{rgb}{0.5,0.5,0.5}
\definecolor{codepurple}{rgb}{0.58,0,0.82}
\definecolor{backcolour}{rgb}{0.95,0.95,0.92}

\newcommand{\datasetFont}{\text}

\newcommand{\ours}{\datasetFont{ConfExplainer}\xspace}

\newcommand{\bamo}{\datasetFont{BA-2motifs}\xspace}
\newcommand{\mutag}{\datasetFont{MUTAG}\xspace}

\newcommand{\ben}{\datasetFont{Benzene}\xspace}
\newcommand{\fluc}{\datasetFont{Fluoride-Carbonyl}\xspace}
\newcommand{\alca}{\datasetFont{Alkane-Carbonyl}\xspace}

\newtheorem{problem}{Problem}

\begin{document}

%%
%% The "title" command has an optional parameter,
%% allowing the author to define a "short title" to be used in page headers.
\title{Is Your Explanation Reliable: Confidence-Aware Explanation on Graph Neural Networks}

%%
%% The "author" command and its associated commands are used to define
%% the authors and their affiliations.
%% Of note is the shared affiliation of the first two authors, and the
%% "authornote" and "authornotemark" commands
%% used to denote shared contribution to the research.
\author{Jiaxing Zhang}
\authornote{Both authors contributed equally to this research.}
\email{jz48@njit.com}
\orcid{0009-0007-8031-661X}
% \author{Xiaoou Liu}
% \authornotemark[1]
% \email{xiaoouli@asu.edu}
\affiliation{%
  \institution{New Jersey Institute of Technology}
  \city{Newark}
  \state{New Jersey}
  \country{USA}
}

\author{Xiaoou Liu}
\authornotemark[1]
\email{xiaoouli@asu.edu}
\affiliation{%
  \institution{Arizona State University}
  \city{Tempe}
  \state{Arizona}
  \country{USA}
}

\author{Dongsheng Luo}
\email{dluo@fiu.edu}
\orcid{0000-0003-4192-0826}
\affiliation{%
  \institution{Florida International University }
  \city{Miami}
  \state{Florida}
  \country{USA}
}

\author{Hua Wei}
\authornote{Corresponding author}
\email{hua.wei@asu.edu}
\orcid{0000-0002-3735-1635}
\affiliation{%
  \institution{Arizona State University}
  \city{Tempe}
  \state{Arizona}
  \country{USA}
}

%%
%% By default, the full list of authors will be used in the page
%% headers. Often, this list is too long, and will overlap
%% other information printed in the page headers. This command allows
%% the author to define a more concise list
%% of authors' names for this purpose.
\renewcommand{\shortauthors}{Zhang et al.}

%%
%% The abstract is a short summary of the work to be presented in the
%% article.
\begin{abstract}
  % Explaining Graph Neural Networks (GNNs) has received increasing attention due to the growing requirement for interoperability, which helps people understand the behaviors of the black box models and gain insights from the model knowledge. However, the explanations generated by the graph explainers are not always reliable. There are many post-hoc instance-level explanation methods, that have been proposed to understand GNN predictions. However, the performance of these methods on the datasets without ground truth are not guaranteed. In this paper, we propose a confidence evaluation module with theoretical support to generate the explainer confidence along with the explanations. The experiment result indicates the advantage of our approach and the effectiveness of the confidence evaluation score.
  Explaining Graph Neural Networks (GNNs) has garnered significant attention due to the need for interpretability, enabling users to understand the behavior of these black-box models better and extract valuable insights from their predictions. While numerous post-hoc instance-level explanation methods have been proposed to interpret GNN predictions, the reliability of these explanations remains uncertain, particularly in the out-of-distribution or unknown test datasets. In this paper, we address this challenge by introducing an explainer framework with the confidence scoring module (\ours), grounded in theoretical principle, which is generalized graph information bottleneck with confidence constraint (GIB-CC), that quantifies the reliability of generated explanations. Experimental results demonstrate the superiority of our approach, highlighting the effectiveness of the confidence score in enhancing the trustworthiness and robustness of GNN explanations.
\end{abstract}


%%
%% Keywords. The author(s) should pick words that accurately describe
%% the work being presented. Separate the keywords with commas.
%% A "teaser" image appears between the author and affiliation
%% information and the body of the document, and typically spans the
%% page.

\received{10 February 2025}
% \received[revised]{10 February 2025}
\received[accepted]{16 May 2025}

%%
%% This command processes the author and affiliation and title
%% information and builds the first part of the formatted document.
\maketitle

\newcommand\kddavailabilityurl{https://doi.org/10.5281/zenodo.15540314}

\ifdefempty{\kddavailabilityurl}{}{
\begingroup\small\noindent\raggedright\textbf{KDD Availability Link:}\\
% please change the following context to include multiple artifacts if necessary.
The source code of this paper has been made publicly available at \url{\kddavailabilityurl}.
\endgroup
}

\input{introduction}
\input{relatedwork}

\input{preliminary}

\input{method}
\input{experiment}
\input{conclusion}

\section*{ACKNOWLEDGMENTS}

The work was partially supported by NSF awards \#2421839, NAIRR \#240120, and NSF No. IIS-2331908. This work used AWS through the CloudBank project, which is supported by National Science Foundation grant \#1925001. The views and conclusions contained in this paper are those of the authors and should not be interpreted as representing any funding agencies. We thank OpenAI for providing us with API credits under the Researcher Access program and Amazon Research Awards.

% \section{Ethics Statement}

% This work is primarily foundational in GNN explainability, focusing on the development of a more
% reliable evaluation algorithm. Its primary aim is to contribute to the academic community by
% enhancing the understanding and implementation of the evaluation process. We do not foresee any
% direct, immediate, or negative societal impacts stemming from the outcomes of our research.

%\section*{References}

\bibliographystyle{ACM-Reference-Format}
\bibliography{sample-base}

%%%%%%%%%%%%%%%%%%%%%%%%%%%%%%%%%%%%%%%%%%%%%%%%%%%%%%%%%%%%

\newpage
\appendix

\input{appendix}

\end{document}

%% file: math_commands.tex
\usepackage{amsmath,amsfonts,bm}

\def\eqref#1{equation~\ref{#1}}
\def\1{\bm{1}}

\def\vx{{\bm{x}}}

\def\mA{{\bm{A}}}

\def\mC{{\bm{C}}}

\def\mM{{\bm{M}}}

\def\mX{{\bm{X}}}

\DeclareMathAlphabet{\mathsfit}{\encodingdefault}{\sfdefault}{m}{sl}
\SetMathAlphabet{\mathsfit}{bold}{\encodingdefault}{\sfdefault}{bx}{n}

\def\gD{{\mathcal{D}}}
\def\gE{{\mathcal{E}}}

\def\gO{{\mathcal{O}}}

\def\gV{{\mathcal{V}}}

\def\sR{{\mathbb{R}}}

\def\emA{{A}}

\def\emC{{C}}

\def\emM{{M}}

\DeclareMathOperator*{\argmin}{arg\,min}

\newcommand{\stitle}[1]{\vspace{1ex}\noindent{\bf #1}}

%% file: introduction.tex
\section{Introduction}
\label{sec:intro}

\begin{figure}[t]
    \centering
    \includegraphics[width=0.4\textwidth]{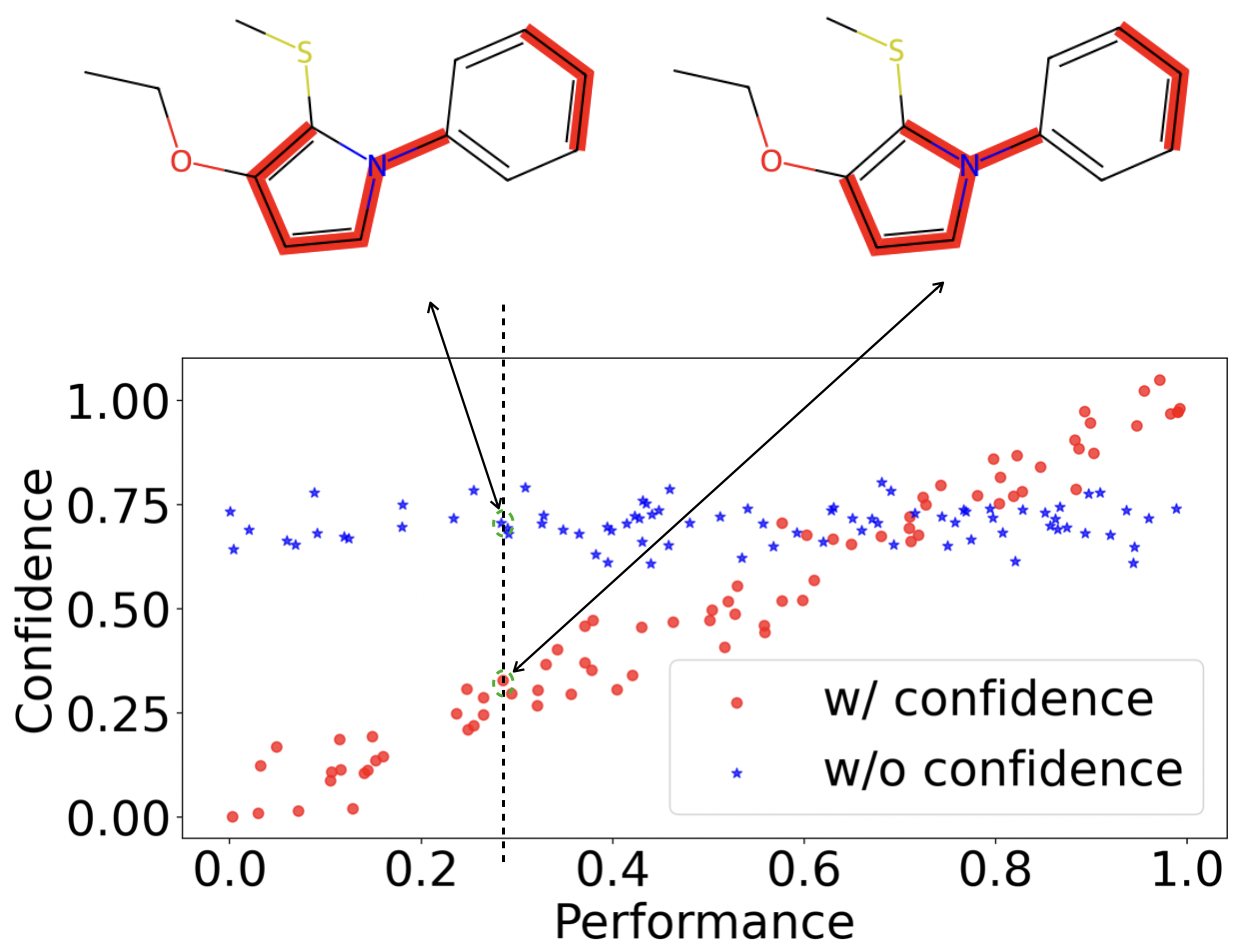}
    % \vspace{-3mm}
    \caption{This figure illustrates the explanations and confidences generated by explainers with and without confidence-aware mechanism. The ground truth explanation in this case is the benzene ring; both samples have a low accuracy.
    % \textcolor{red}{decrease the front size and modify the molecular examples}
    % The red points demote the explanations generated by the confidence-aware explainer; The blue points demote the explanations generated by the non-confidence-aware explainer. As shown in the figure, the molecular explanation on the right has a high performance aligned with confidence, while the left explanation has a lower accuracy, despite similar confidence. 
    }
    % \vspace{-0.3cm}
    \label{fig:conf_tuition_demo}
\end{figure}

Graph Neural Networks (GNNs) have emerged as powerful tools for learning from graph-structured data, with applications spanning recommendation systems~\cite{fan19social, min21social}, molecular chemistry~\cite{ronast2024neurips3dmole, xufeng2024kdd3dgraphx}, traffic prediction~\cite{wang20traffic, Li_Zhu_2021_traffic,wu19graphwave,ijcai2018p0505}, and social network analysis~\cite{zhou2020graphreview}. Despite their impressive predictive capabilities, GNNs lack transparency, making it challenging to understand their decision-making process. This opacity is particularly concerning in high-stakes applications such as healthcare, finance, and scientific discovery, where understanding model predictions is as critical as the predictions themselves~\cite{abdar2021uqreview}. Consequently, there is a growing interest in explainability methods that aim to elucidate how GNNs arrive at specific predictions~\cite{dai2024explainersurvey}.
A prominent direction for post-hoc explainability in GNNs involves identifying critical substructures (i.e., nodes and edges) that most influence the model’s decisions. Methods such as GNNExplainer\cite{ying2019gnnexplainer} and PGExplainer\cite{luo2020parameterized} generate explanations by masking or perturbing parts of the input graph to determine which components are most influential. More recently, Graph Information Bottleneck (GIB)-based methods~\cite{wu2020graph,miao2022interpretable,yu2020graph,chen2024generating} have provided a principled framework for explainability, aiming to maximize mutual information between the explanation and the target output while compressing redundant information. The Information Bottleneck (IB) principle has demonstrated significant potential in balancing between representation relevance and compression, making it a robust theoretical foundation for explainable GNNs.

%sec2
% However, existing IB-based approaches largely focus on explaining node-level or graph-level predictions without explicitly addressing the confidence of explanations or handling uncertainty in the data. This gap becomes particularly pronounced when dealing with OOD data or when the quality of explanations varies across different parts of the graph. It will also lead to a mismatch between the explanation performance and the explainer's confidence during the explanation, where we assume that the performance of the explanation should align with the confidence. As the intuition example shown in Fig.~{\ref{fig:conf_tuition_demo}}, the red points represent the explanations generated by the confidence-aware explainer; while the blue points represent the explanations generated by the explainer without a confidence mechanism. There are pair of explanations retrieved with similar low accuracy. However, the non-confidence-aware approach couldn't inform the user whether or not the explanation is reliable, while the confidence-aware approach could give a proper confidence estimation.

However, existing GIB-based methods and other post-hoc explainability techniques largely overlook an essential aspect--the confidence of the explanation itself. While these methods generate explanations, they do not quantify how reliable or trustworthy those explanations are~\cite{wang2024uncertaintygnn}. This limitation is particularly concerning when models are deployed in out-of-distribution (OOD) or unknown test set scenarios, where explanations may degrade unpredictably. An explanation with high confidence but poor fidelity can be misleading~\cite{dexai2024trustgnn}, potentially leading to incorrect human interpretations in critical decision-making settings (e.g., medical diagnosis or financial risk assessment~\cite{li2020graphmedical,zhang2024gnnfinancial}).

We hypothesize that the confidence of an explainer should align with its performance. If an explainer produces unreliable explanations, it should reflect low confidence, rather than misleadingly indicating high certainty. Conversely, high-quality explanations should be accompanied by high confidence scores. However, existing explainability methods do not provide such confidence quantification, leading to potential mismatches between explanation quality and user trust.
As illustrated in Figure~\ref{fig:conf_tuition_demo}, the confidences of explanations generated by non-confidence-aware explainers (blue points) fail to distinguish between high- and low-fidelity explanations, whereas our confidence-aware approach (red points) provides well-calibrated confidence scores that align with the true reliability of the explanation. This highlights the necessity of confidence-aware explanation mechanisms in GNNs.

%sec3
%Drawing inspiration from recent advancements in incorporating confidence-aware mechanisms into the IB framework, we propose to extend the IB principle to graph neural networks by introducing a confidence evaluation module. This module dynamically evaluates the reliability of graph-level explanations, allowing for more robust and interpretable insights. By leveraging a confidence-aware encoder within the graph information bottleneck framework, our method will: 1. Quantify and incorporate confidence into the explanation process, enabling users to assess the reliability of generated insights. 2. Distinguish between regions of the graph with high and low confidence, facilitating targeted improvements in both explanations and predictions. 3. Enhance the model's ability to generalize to OOD data by explicitly modeling uncertainty in explanations.

To address this limitation, we propose a Generalized Graph Information Bottleneck with Confidence Constraints (GIB-CC), an extension of GIB that explicitly integrates confidence estimation into the explanation generation process and a confidence-aware explainer framework accordingly. 
We conduct extensive experiments to evaluate the effectiveness of our confidence evaluation module. Our results demonstrate that incorporating confidence scores not only improves the performance of GNN explainers but also provides valuable insights into the reliability of their explanations. This advancement holds significant promise for deploying GNNs in critical applications where understanding model decisions is as important as the decisions themselves. In summary, this work aims to bridge the gap between interpretability and reliability in GNN explanations, targeting critical challenges in applying GNNs to real-world scenarios.
Our approach aligns with the growing demand for trustworthy AI, where understanding not only what a model predicts but also how confident it is in its explanations is essential for responsible decision-making.
Our key contributions include: 

1. {Confidence-Aware Explainer (\ours)}, where we introduce an explainer with confidence scoring model that dynamically assesses the reliability of graph-level explanations. This enables users to gauge whether an explanation can be trusted or should be interpreted with caution. 

2. {Confidence-Guided Information Bottleneck Objective}: We extend the GIB framework by incorporating confidence constraints (GIB-CC), allowing our framework to adaptively weigh explanations based on their reliability. This approach effectively reduces uncertainty and enhances the robustness of the explanations, particularly in OOD or unknown test set scenarios. 

3. {Empirical Validation on Diverse Datasets}: We conduct comprehensive experiments on multiple benchmark datasets, including synthetic and molecular graphs, demonstrating our framework's performance in both explanation fidelity and confidence calibration.

%% sec4
 
% Significance and Impact By introducing confidence-aware explanations for GNNs, our work addresses a critical gap in existing explainability research. Our framework enables: Better trust calibration, ensuring that high-confidence explanations are indeed reliable. Improved robustness, particularly in cases where the explainer faces OOD data. More responsible AI deployment, where decision-makers can assess both what a model predicts and how confident it is in its explanation. This advancement is crucial for the broader goal of trustworthy AI, aligning with the increasing need for interpretable and accountable machine learning models.

%\vspace{-0.5cm}

%% file: relatedwork.tex
\section{Related Work}
\label{sec:relatedwork}

%\subsection{Graph Neural Networks}
% The use of graph neural networks (GNNs) is on the rise for analyzing graph structure data, as seen in recent research studies~\cite{dai2022towards, fan19social, hamilton2017inductive}. There are two main types of GNNs: spectral-based approaches~\cite{bruna2013spectral, kipf2016semi, tang2019chebnet} and spatial-based approaches~\cite{atwood2016diffusion, duvenaud2015convolutional, xiao2021learning}. Despite the differences, message passing is a common framework for both, using pattern extraction and message interaction between layers to update node embeddings. However, GNNs are still considered a black box model with a hard-to-understand mechanism, particularly for graph data, which is harder to interpret compared to image data. To fully utilize GNNs, especially in high-risk applications, it is crucial to develop methods for understanding how they work.

\textbf{GNN Explanation.}
Many attempts have been made to interpret GNN models and explain their predictions~\cite{rgexp, ying2019gnnexplainer, luo2020parameterized, subgraphx, spinelli2022meta, wang2021causal}. These methods can be grouped into two categories based on granularity: (1) instance-level explanation, which explains the prediction for each instance by identifying significant substructures~\cite{ying2019gnnexplainer, subgraphx, rgexp, zhang2024llmexplainerlargelanguagemodel}, and (2) model-level explanation, which seeks to understand the global decision rules captured by the GNN~\cite{luo2020parameterized, spinelli2022meta, bald19expgcn}. From a methodological perspective, existing methods can be classified as (1) self-explainable GNNs~\cite{bald19expgcn, dai21towards}, where the GNN can provide both predictions and explanations and (2) post-hoc explanations~\cite{ying2019gnnexplainer, luo2020parameterized, subgraphx}, which use another model or strategy to explain the target GNN. In this work, we focus on post-hoc instance-level explanations, which involve identifying instance-wise critical substructures to explain the prediction. Various strategies have been explored, including gradient signals, perturbed predictions, and decomposition.

Perturbed prediction-based methods are the most widely used in post-hoc instance-level explanations. The idea is to learn a perturbation mask that filters out non-important connections (label irrelevant) and identifies dominant substructures (label preserving) while preserving the original predictions. For example, GNNExplainer~\cite{ying2019gnnexplainer} uses end-to-end learned soft masks on node attributes and graph structures, while PGExplainer~\cite{luo2020parameterized} incorporates a graph weight generator to incorporate global information. RG-Explainer~\cite{rgexp} uses reinforcement learning technology with starting point selection to find important substructures for the explanation.

\noindent\textbf{Confidence Estimation.}
However, existing GIB-based methods lack confidence estimation, which is essential for distinguishing between reliable and misleading explanations, particularly under distribution shifts. Although techniques such as Monte Carlo Dropout and Deep Ensembles~\cite{wang2024uncertaintygnn} have been applied to uncertainty estimation in GNNs, they focus on predictive confidence rather than explanation confidence, leaving a gap in quantifying the trustworthiness of explanations themselves.

To address this limitation, we propose Generalized Graph Information Bottleneck with Confidence Constraints (GIB-CC), a novel framework that integrates confidence into post-hoc GNN explainability. Our method dynamically quantifies explanation reliability by learning a confidence-aware representation, ensuring that explanations align with their true fidelity. Unlike ensemble-based approaches, our confidence estimation is self-contained, computationally efficient, and theoretically grounded, making it applicable to real-world scenarios where understanding both what a model predicts and how confident it is in its explanations is critical.

%% file: preliminary.tex
\section{Preliminary}

\subsection{Notations and Problem Definition}
% \subsubsection{Introduce the method}
We denote a graph as $G= (\mathcal{V}, \mathcal{E}; \mX, \mA)$, where $\mathcal{V} = \{v_1, v_2, ..., v_n\}$ represents a set of $n$ nodes and $ \mathcal{E} \in \mathcal{V} \times \mathcal{V}$ represents the edge set. Each graph has a feature matrix $ \mX \in \sR^{n\times d} $ for the nodes. where in $\mX $, $ \vx_i  \in \sR^{1\times d} $ is the $d$-dimensional node feature of node $v_i$. $\mathcal{E}$ is described by an adjacency matrix $ \mA \in \{0,1\}^{n\times n}$. $\emA_{ij} = 1$ means that there is an edge between node $v_i$ and $v_j$; otherwise, $\emA_{ij} = 0$. 

For graph classification tasks, each graph $G_i$ has a label $Y_i \in \mathcal{C}$, with a GNN model $f$ trained to classify $G_i$ into its class, i.e., $f:(\mX, \mA) \mapsto \{1, 2, ..., C\}$.

\begin{problem} [Post-hoc Instance-level GNN Explanation]
\label{prob:exp}
Given a trained GNN model $f$, for an arbitrary input graph $G=(\mathcal{V}, \mathcal{E}; \mX, \mA)$, the goal of post-hoc instance-level GNN explanation is to find a sub-graph $G^{*}$ that can explain the prediction of $f$ on $G$. 
\end{problem}

Informative feature selection has been well studied in non-graph structured data~\cite{Li17featureselect},  and traditional methods, such as concrete autoencoder~\cite{balin2019concrete}, can be directly extended to explain features in GNNs. In this paper, we focus on discovering important typologies and their confidence. Formally, the obtained explanation $G^{*}$ is depicted by a binary mask $\mM \in \{0, 1\}^{n\times n}$ on the adjacency matrix $\mA$, e.g., $G^{*} = ( \gV, \gE, \mA\odot \mM; \mX)$, $\odot$ means elements-wise multiplication. The mask highlights components of the original graph $G$ that are essential for $f$ to make the prediction. More importantly, we would like to obtain the confidence of explanations given $\mM$ and $G$:

\begin{problem} [Confidence-Aware Explanation]
\label{prob:conf}
Given an Explainer model $E$, for an arbitrary input graph $G= (\mathcal{V}, \mathcal{E}; \mX, \mA)$ and explanation $G^*$, the goal of confidence scoring model $f_\mC$ is giving the confidence score matrix $\mC=\{\emC_{ij} | \emC_{ij} \in [0, 1]\} \in \mathbb{R}^{n\times n}$ that can reflect the confidence of $E$ on edges in $G^*$. 
\end{problem}

%Our methodology is designed to address problem~\ref{prob:conf}.
% \subsubsection{GNN Explanation}

% We denote an explanation as an edge mask $M_i$ for graph $G_i$, where $M_i = [w_1, w_2, ..., w_e]$, $e$ means the number if the edges in graph $G_i$.

%% file: method.tex
\begin{figure*}
    \centering
    \includegraphics[width=1\textwidth]{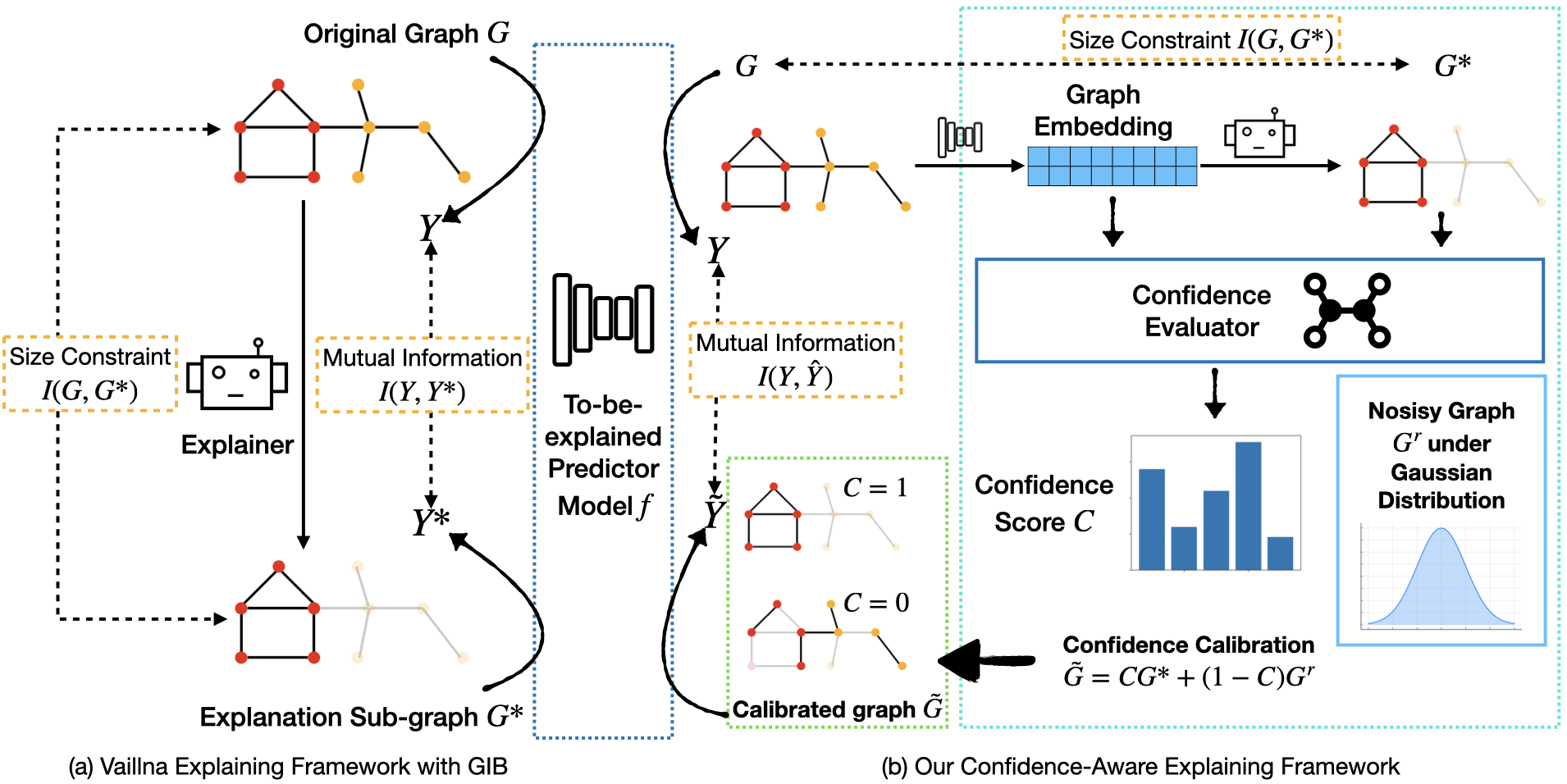}
    % \vspace{-3mm}
    \caption{This figure illustrates the difference between our approach and previous approaches, based on GIB. Figure (a) on the left shows the explaining framework optimized by Graph Information Bottleneck. Figure (b) on the right shows our confidence-aware explaining framework. We first generate the explanation sub-graph $G^*$ from the original graph $G$ with the explainer model. Before confidence evaluation, the mask vector of edges in $G^*$ would be concatenated with the hidden state embedding of the original graph $G$. After generating the confidence score $C$, we aggregate $G^*$ with a random  Gaussian noisy graph $G^r$, weight by $C$ and $1-C$, as $\tilde G$. Finally, we compute the confidence loss and GIB loss with $C$, $G^*$, and $\tilde G$.}
    % \vspace{-3mm}
    \label{fig:main}
\end{figure*}

\section{Methodology}
\label{sec:method}
% In this section, we investigate the Graph Information Bottleneck (GIB) with confidence constraints and propose an efficient confidence model to incorporate and estimate the explainer's confidence during explaining, driven by theoretical analysis.

The original Graph Information Bottleneck (GIB) framework provides an effective paradigm for extracting explanatory sub-graphs from a given graph $G$. However, existing GIB-based explainability methods lack an explicit mechanism for measuring the confidence of explanations. To address this, we introduce the Generalized Graph Information Bottleneck with Confidence Constraints (GIB-CC), which extends GIB by integrating a confidence-aware mechanism into the optimization objective.

\subsection{Formalization of GIB-CC}

Given a graph $G \sim P_\mathcal{G}$ with its label $Y \sim P_\mathcal{Y}$, and it's sub-graph $G' \subseteq G$, the Graph Information Bottleneck, aiming to find the optimized explanation $G^*$, objective is formulated as Eq.~(\ref{eq:gib}):
\begin{equation}
    \label{eq:gib}
    G^* = \underset{G'}{\argmin}\quad I(G, G') - \alpha I(Y,G'),
\end{equation}
% where $G^*$ is the optimized sub-graph explanation, $G'$ is candidate sub-graph explanation, and $Y$ is the ground truth target variable for the original graph $G$. However, this objective function lacks the ability to measure the explainers' confidence during explaining.
where $G^*$ represents the optimized explanation sub-graph, and $G'$ denotes a candidate sub-graph. The goal is to minimize structural constraints between $G$ and $G'$ while retaining mutual information relevant to $Y$. However, this objective lacks an explicit mechanism for estimating the confidence of the explanations.

% \subsection{Generalized Graph Information Bottleneck with Confidence Constraints (GIB-CC)}

%We introduce our generalized Graph Information Bottleneck with the Confidence constraints. Formally, it's written as :

\subsubsection{Confidence-aware GIB} To estimate the confidence of the explanations, we introduce a confidence matrix $\mC$, which serves as a measure of the reliability of each edge in the sub-graph. We reformulate the objective with the confidence constraints as follows:
\begin{equation}
    \label{eq:gibcc}
    \begin{aligned}
     \argmin_{G'}  &  {\quad I(G, G') - \alpha I(Y, \tilde G)}, \\
    & \text{where } \tilde G = \phi(G', \mC) = \mC \odot G' + (\mathbb{1}-\mC) \odot G^r, \\
    & \text{s.t. } I(\mC, G^r;Y | G')=0.
    \end{aligned}
\end{equation}
% where $\mC$ is the confidence matrix for the sub-graph $G^*$, and $G^r = ( \gV, \gE, \mA\odot \mathcal{N}(0, 1); \mX)$ \jx{should noise add to $G$ or the fully connected graph?} represents graph-level Gaussian noise.
Here, $\tilde G$ is a calibrated version of the explanation sub-graph, incorporating Gaussian noise $G^r: \mA\odot \mM^r$ weighted by the confidence score $\mC$, where each entry $\emM^r_{ij}$ in $\mM^r$ is randomly sampled from the Gaussian distribution. The term $\mC \odot G^r$ represents the confidence-weighted explanation over the 
% \hua{need to specify what $\mC$ is multiplying with - is it $A$, or $M$?}, 
while $(\mathbb{1}-\mC) \odot G^r$ introduces controlled noise to calibrate uncertainty. $\mathbb{1}$ is a ones matrix with same shape as $\mA$, $\mM$, and $\mC$.
% \hua{what is \text{D}?} 
The constraints ensure that the calibrated graph retains the same level of dependency on the original graph and labels as the original explanation. 
%\hua{define C and add notation table}

\subsubsection{Mutual Information Interpretation.} To further justify the role of confidence in our objective, we decompose the mutual information terms:
\begin{equation} I(Y, \tilde{G}) = H(Y) - H(Y | \tilde{G}), \end{equation}
where the conditional entropy can be rewritten as:
\begin{equation} H(Y|\tilde{G}) = H(Y | G') + I(\mC, G^r; Y | G'). \end{equation}
% By assuming $I(\mC, G^r; Y|G')=0$ 
% \hua{need explanation why this assumption is realistic}
Given $G^r$ as an independent sampled graph noise and $\mC$ as an extrinsic generated score, we have $I(\mC, G^r; Y|G')=0$. A detailed illustration could be found in Appendix~{\ref{sec:justification_core_condition}}. Therefore, we obtain:
\begin{equation} H(Y|\tilde{G}) = H(Y | G'), \end{equation}
which ensures that our framework, with the confidence-aware formulation, maintains equivalence with the original GIB formulation. This guarantees that adding confidence constraints does not distort the explanation quality while enabling confidence estimation. Therefore, we have property (1) as follows:

\stitle{Property 1.} \emph{ The confidence-aware GIB objective, Eq. (\ref{eq:gibcc}) is equivalent to vanilla GIB, Eq. (\ref{eq:gib}).}

This property can be proved by considering the conditions of $I(\mC, G^r;Y | G')=0$, then we have:
$$H(Y|\tilde G) = H(Y|G^*) + I(\mC, G^r; Y|G^*) = H(Y|G^*; \mC, G^r).$$
Thus, the optimal solutions of GIB and our confidence-aware version are equivalent. 
The advantage of our objective is that by introducing the confidence matrix $\mC$ that weighted with the Gaussian noise $G^r$, we can generate the explanation sub-graph $G^*$ with a confidence $\mC$ with stochasticity. %This stochasticity could alleviate the distribution shifting issue which is well-studied in existing work~\cite{zhang2023mixupexplainer,zhang2023regexplainer}. 
Following exiting work~\cite{ying2019gnnexplainer,luo2020parameterized,zhang2023regexplainer}, we can further approximate $H(Y|G^*;\mC, G^r)$ with $\text{CE}(Y, \tilde Y)$ in classification tasks or $\text{MSE}(Y, \tilde Y)$ in regression tasks, where $\tilde Y = f \circ \phi(G^*, \mC)$ is the predicted label of $ \tilde G$ made by the model $f$. Especially when $\mC$ is a one's vector, our objective degenerates to the vanilla approximation. Formally, the design of the confidence evaluation model and the loss function derived from the generalized GIB-CC objective for GNN explanation are described in the following sections.
% \begin{equation}
% \begin{aligned}
% \label{eq:gib-ours}
%      \argmin_{\mC, G^*} & { \quad I(G, G^*)+\alpha \text{CE}(Y, \tilde Y)}\\
%     & \text{s.t. } \text{D}(G, \tilde G) =0 , I(\mC, G^r;Y | G^*)=0.
% \end{aligned}
% \end{equation}

\subsection{Confidence Evaluation Framework}
In this section, we propose our confidence-aware explaining framework based on the theoretical analysis previously. 

\subsubsection{Confidence Scoring Model. }
%The confidence score $C_{G^*} = \text{RELU}(\text{MLP}(G, G^*))$, for each edge in $G^*$, $c_i \in [0, +\inf]$. Activate function $A(x) = \frac{1}{e^{-x}}$, where $x \ in [0, +\inf]$ and $A(x) \in [0, 1]$.
To quantify the reliability of explanations, we introduce a confidence scoring mechanism based on a learnable function $f_\mC$, which generates a confidence score for each edge in the explanatory sub-graph $G'$. The confidence matrix is computed with an MLP and an activation layer as follows:
\begin{equation} \mC = \text{RELU}(\text{MLP}(f_\text{emd}(G), \mM)), \end{equation}
where MLP is a multi-layer perceptron that takes the graph representation embeddings of the original graph $G$ and its explanation mask $\mM$. $f_\text{emd}$ is the encoder part of the to-be-explained GNN model, where we extract the embedding before predicting. The activation function ensures that the confidence scores remain non-negative.

For each edge in $G^*$, the confidence score is given by $\emC_{ij} = \frac{1}{1 + e^{-k_{ij}}}$, where $k_{ij}$ is the raw confidence value from the MLP. This sigmoid activation ensures that confidence scores lie in the range $(0, 1)$, making them interpretable as probability estimates.

\subsubsection{Loss Function. } 
% Follow the previous work~\cite{ying2019gnnexplainer, luo2020parameterized, zhang2023mixupexplainer}, We train the explainer with GIB loss, which is formally written as:
% \begin{equation}
% \label{eq:gibloss}
%     \mathcal{L}_\text{GIB} = \mathcal{L}_{\text{size}}(G, G^*) + \alpha \text{CE}(Y, f(\tilde G)).
% \end{equation}
To optimize the explainer and the confidence estimator, we define a joint objective function consisting of Graph Information Bottleneck (GIB) Loss and Confidence Loss.

\stitle{GIB Loss. } 
%We train the confidence evaluator with a confidence loss, which is designed with the following principle: \textbf{Given an edge prediction possibility of the explanation sub-graph $G^*$, while the edge classification is correct, the closer it's to the ground truth, the higher confidence it should have; The farther it's to the ground truth, the lower confidence it should have. On the other side, the closer it's to the ground truth, the lower confidence it should have; The farther it's to the ground truth, the higher confidence it should have. \jx{It may not be correct}} Based on the idea, we first design a true prediction mask:
% \begin{equation}
%     \begin{aligned}
%         & M_\text{tp}(\tilde y, y) = 1 \text{ if } \tilde y \text{ is correctly classified }, \\
%         & M_\text{tp}(\tilde y, y) = 0 \text{ if } \tilde y \text{ is not correctly classified},
%     \end{aligned}
% \end{equation}
% where $\tilde y$ is the prediction label of edge and $y$ is the ground truth label of edge. Therefore, we formally write our confidence loss as:
% \begin{equation}
% \label{eq:lossconf}
%     \mathcal{L}_C = \beta \sum_{i=1}^{N} \frac{(C_i * (\tilde{y}_i - y)^2)}{N},
% \end{equation}
% where hyper-parameter $\beta$ is the weight of confidence loss.
Following prior work~\cite{ying2019gnnexplainer, luo2020parameterized}, the overall loss function including the GIB loss, where the GIB loss is formulated as:

\begin{equation} \mathcal{L}_\text{GIB} = \mathcal{L}_{\text{size}}(G, G^*) + \alpha \text{CE}(Y, f(\tilde G)). \end{equation}

The first term $\mathcal{L}_{\text{size}}(G, G')$ ensures that the sub-graph remains compact, while the second term minimizes cross-entropy (CE) loss between the predicted label $\tilde Y$ and target label $Y$.

\stitle{Confidence Loss.} We design the confidence loss based on the following principle:
Given an edge prediction probability in $G'$, if the classification is correct, the confidence should be high when close to the ground truth and low when far from the ground truth. Conversely, for incorrect predictions, confidence should be low when it is near the ground truth and high when it is far from it.
Based on this principle, we first define a true prediction mask:

\begin{equation} 
M_\text{tp}(\tilde y, y) = 
\begin{cases} 
1, & \text{if } \tilde{y} \text{ is correctly classified}, \\ -1, 
& \text{otherwise}. 
\end{cases} 
\end{equation}

Using this mask, our final confidence loss is written as:

\begin{equation} \mathcal{L}_C = \beta \sum M_{\text{tp},i} \frac{(C_i * (\tilde{y}_i - y)^2)}{N}, \end{equation}

where $\beta$ is a hyperparameter controlling confidence regularization. This encourages the model to assign high confidence to correct explanations and low confidence to uncertain ones.

\stitle{Final Objective.} Combining both terms, our final optimization objective is:
\begin{equation} \mathcal{L} = \mathcal{L}_\text{GIB} + \lambda \mathcal{L}_C, \end{equation}
where $\lambda$ balances explanation quality and confidence estimation.

\stitle{Training Procedure. } Since the confidence loss contains both confidence score and explanation mask, to avoid of trivial solution, we train the explainer model and confidence evaluator model iteratively. When training one of them, another one is frozen. Our training algorithm is shown in Appendix~{\ref{sec:algo}}.

\subsection{Time Complexity Analysis}
% In this section, we analyze the time complexity of our confidence evaluation module.
Given the hidden state embedding $e_G$ with shape of $(d_e, d_n)$ of an original graph $G$, sub-graph explanation $G^*$ with shape of $(\gD_e, 1)$, and a confidence model $f_\mathcal{C}$ with fully connection layer which map dimension $\gD_m$ to $1$, the complexity of concatenate of the $e_g$ and $G^*$ is $\gO(|\gD_e|$, where $|\gD_e|$ denotes the number of edges or dimension of $e_G$ of the original graph $G$.
And the complexity of generating the confidence score is $\gO(|\gD_e|*(|\gD_n|+1)*1)$. For generating the confidence score, the complexity is $\gO(|\gD_e|*(|\gE_n|+2))$. Considering the number $2$ as a small constant and $\gD_n$ is treated as a hyper-parameter with a constant value, the overall complexity of our confidence generation is $\gO(|\gD_e|)$.

%% file: experiment.tex
\section{Experimental Study}
We conduct comprehensive experimental studies on benchmark datasets to empirically verify the effectiveness of the proposed \ours. Specifically, we aim to answer the following research questions:
\noindent~$\bullet$  RQ1: How does the proposed method perform compared to the baseline methods?
\noindent~$\bullet$  RQ2: Can the proposed framework generate better confidence scores?
\noindent~$\bullet$  RQ3: How do the confidence-aware explanations look like? Is it visually correct?

In addition to research questions, we conduct the experiments, including training time cost comparison, ablation study, and hyperparameter tuning, which are put in the appendix.

\input{tabletexs/table1_rq1}

\begin{figure*}[t!]
% \vspace{-1cm}
    \centering
    \scalebox{0.95}{
    \begin{tabular}{ccccc}
        \subfigure{\includegraphics[width=0.19\textwidth]{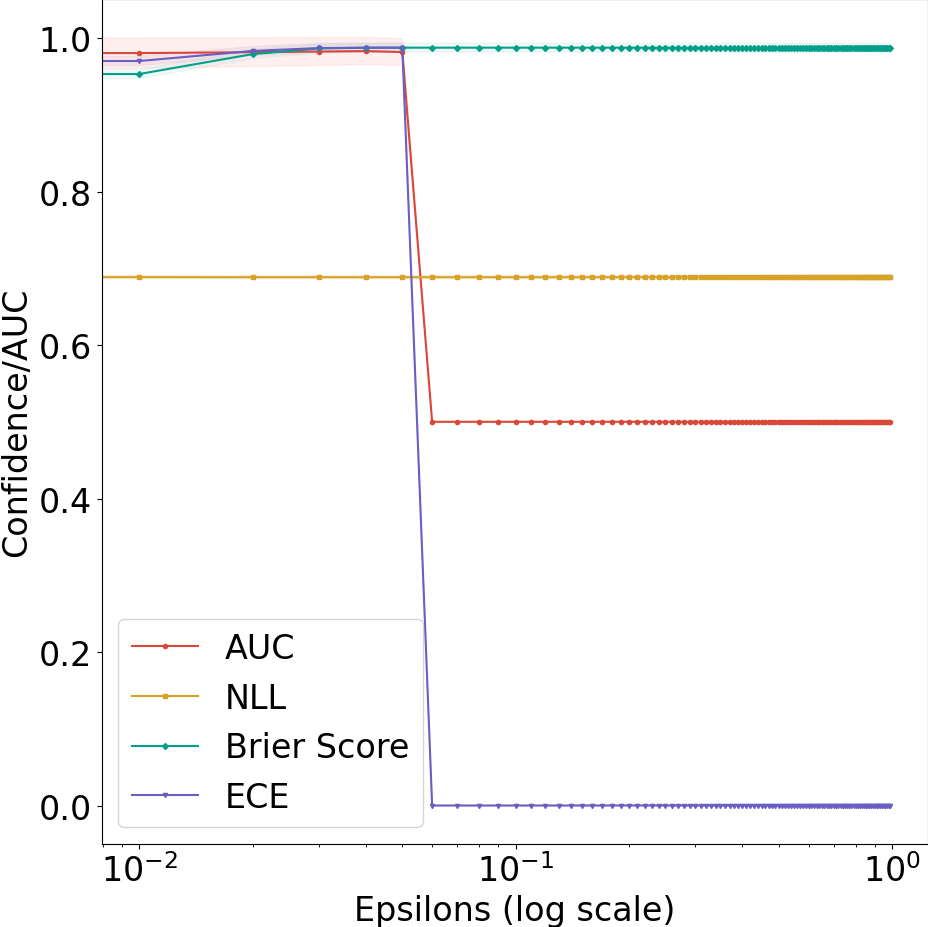}}  & 
        \subfigure{\includegraphics[width=0.19\textwidth]{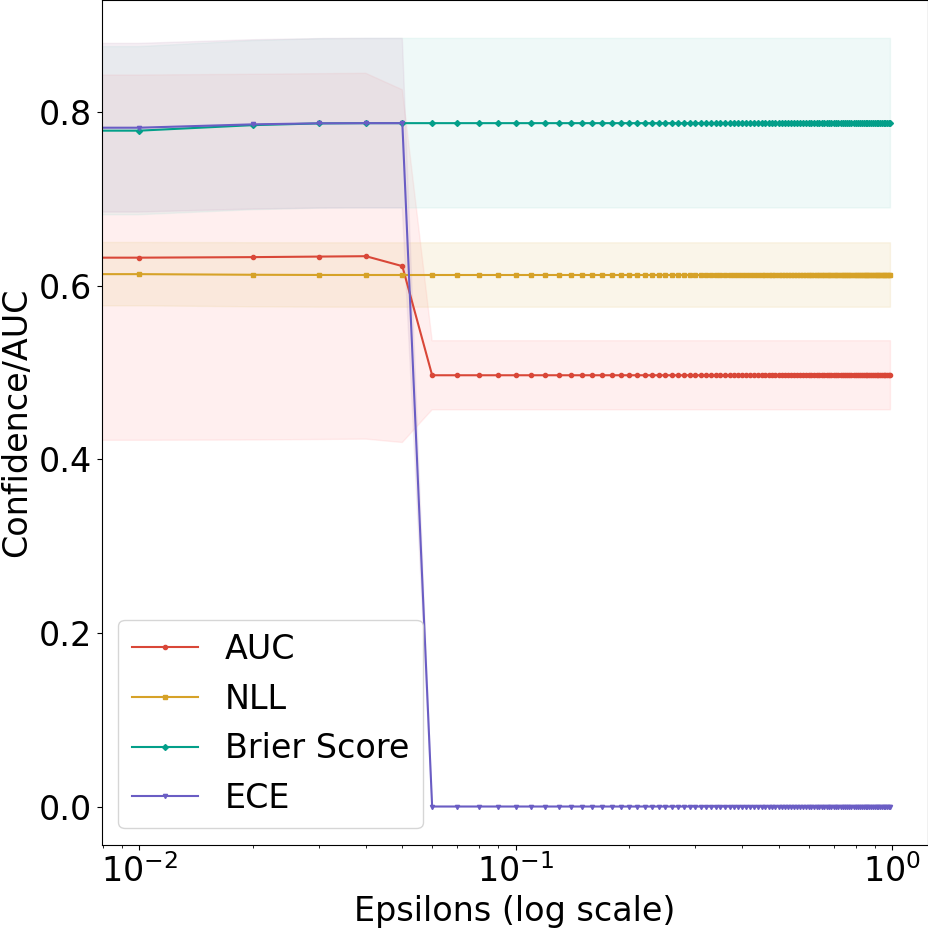}}   & 
        \subfigure{\includegraphics[width=0.19\textwidth]{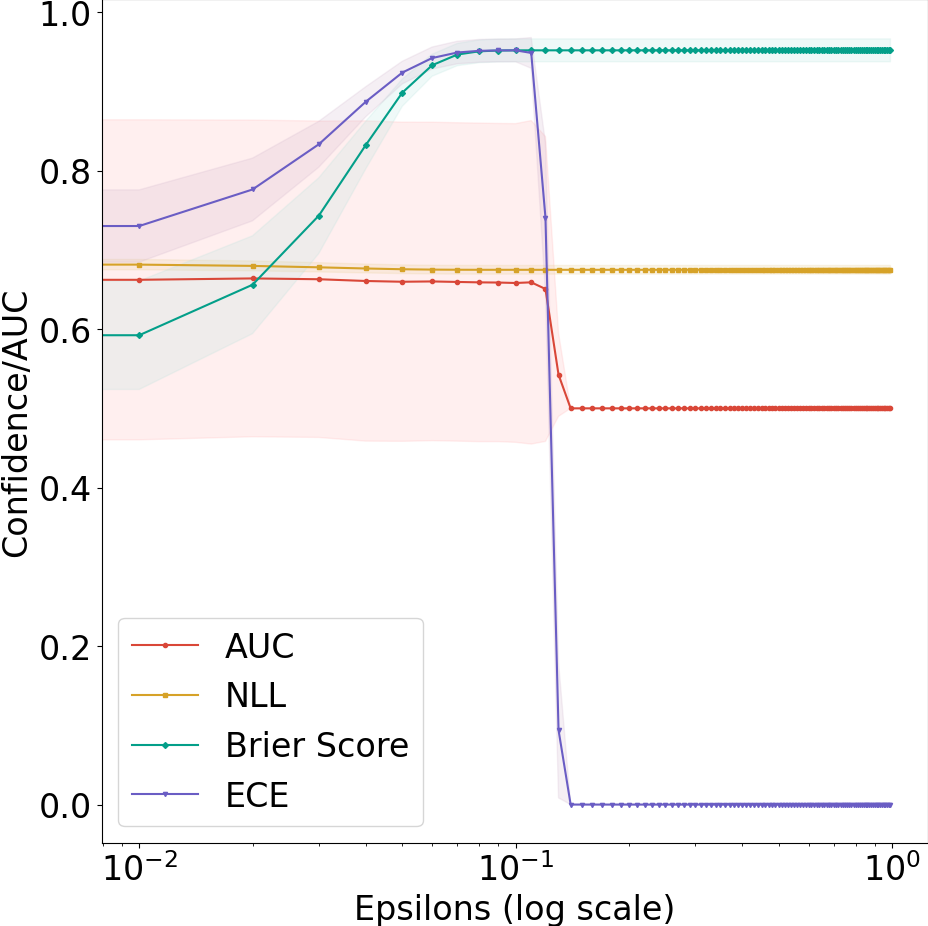}}  & 
        \subfigure{\includegraphics[width=0.19\textwidth]{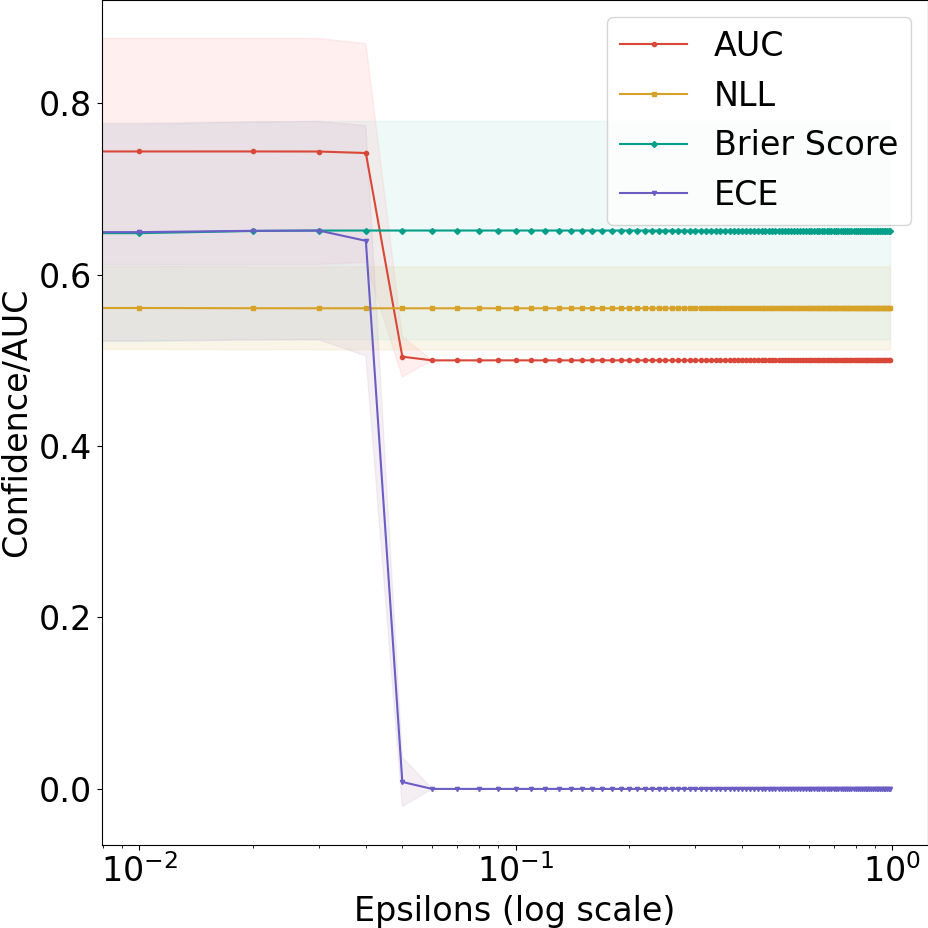}} & 
        \subfigure{\includegraphics[width=0.19\textwidth]{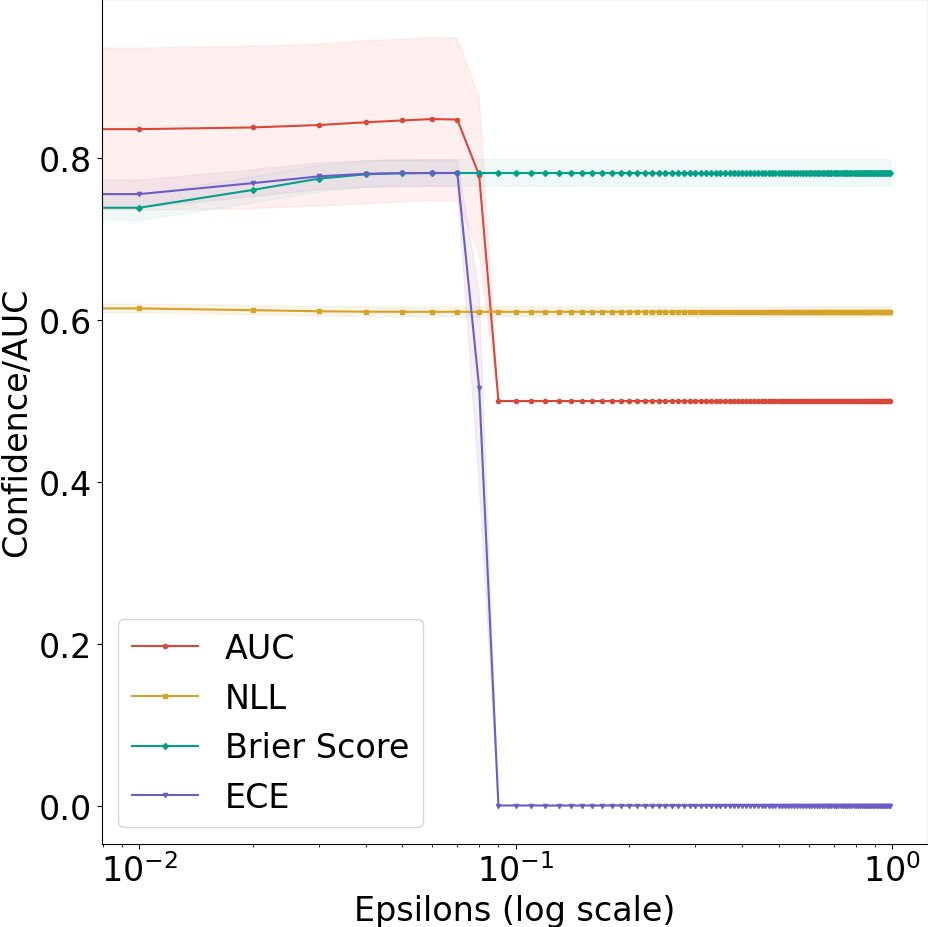}} \\
        \vspace{5pt}
        \subfigure{\includegraphics[width=0.19\textwidth]{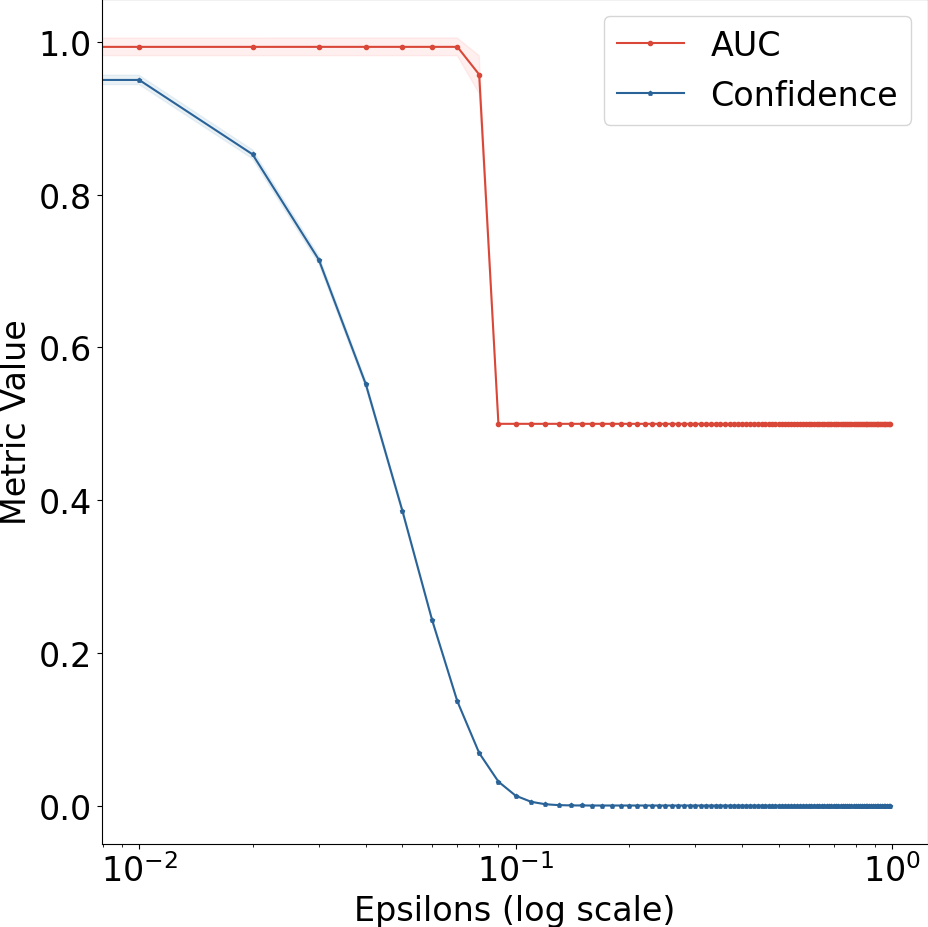}} &
        \subfigure{\includegraphics[width=0.19\textwidth]{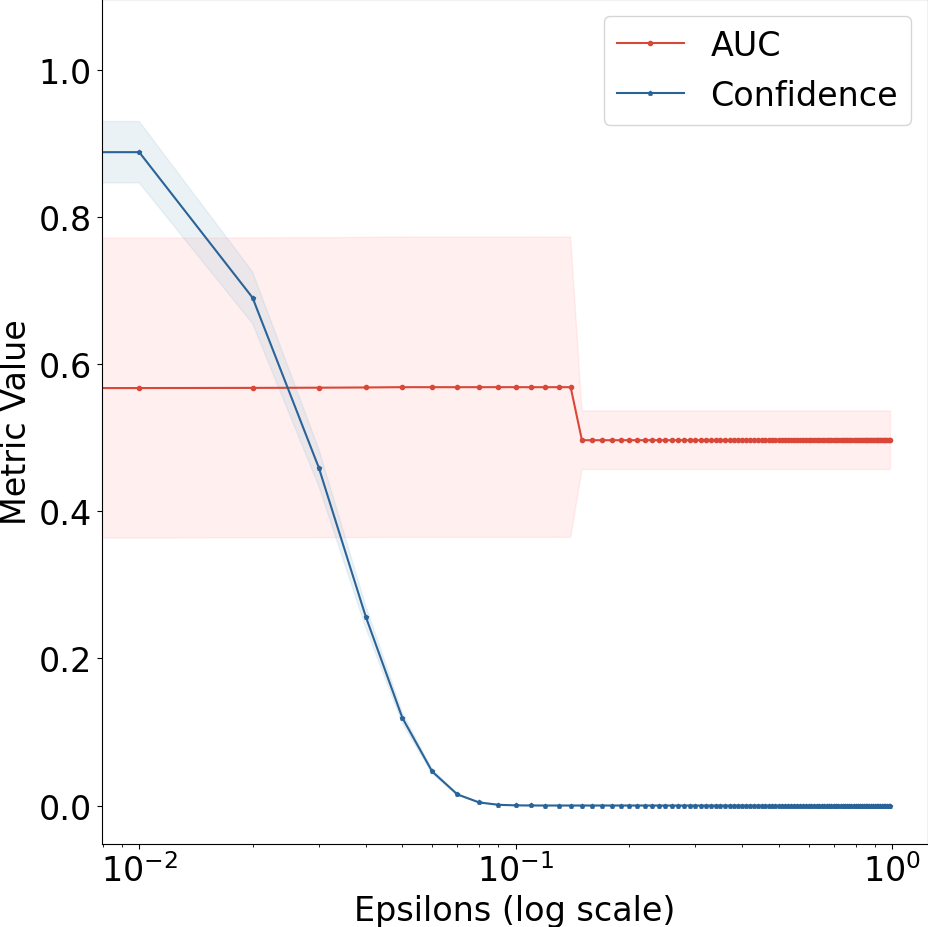}} &
        \subfigure{\includegraphics[width=0.19\textwidth]{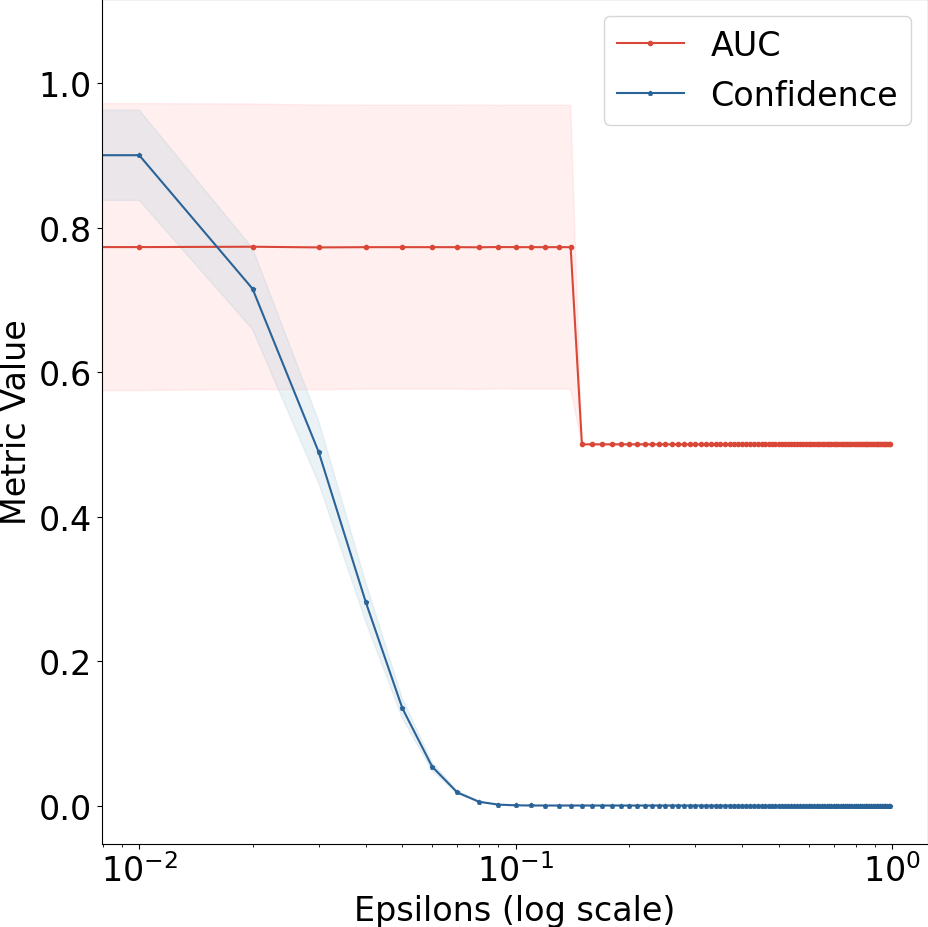}} &
        \subfigure{\includegraphics[width=0.19\textwidth]{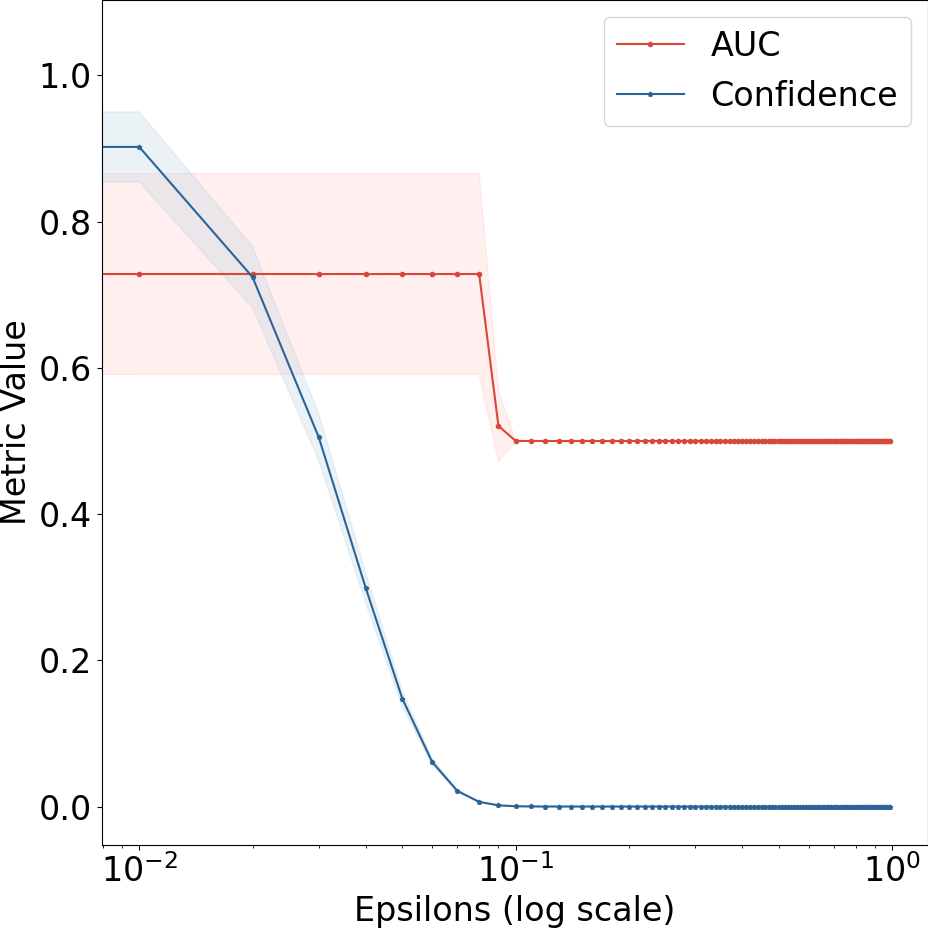}} &
        \subfigure{\includegraphics[width=0.19\textwidth]{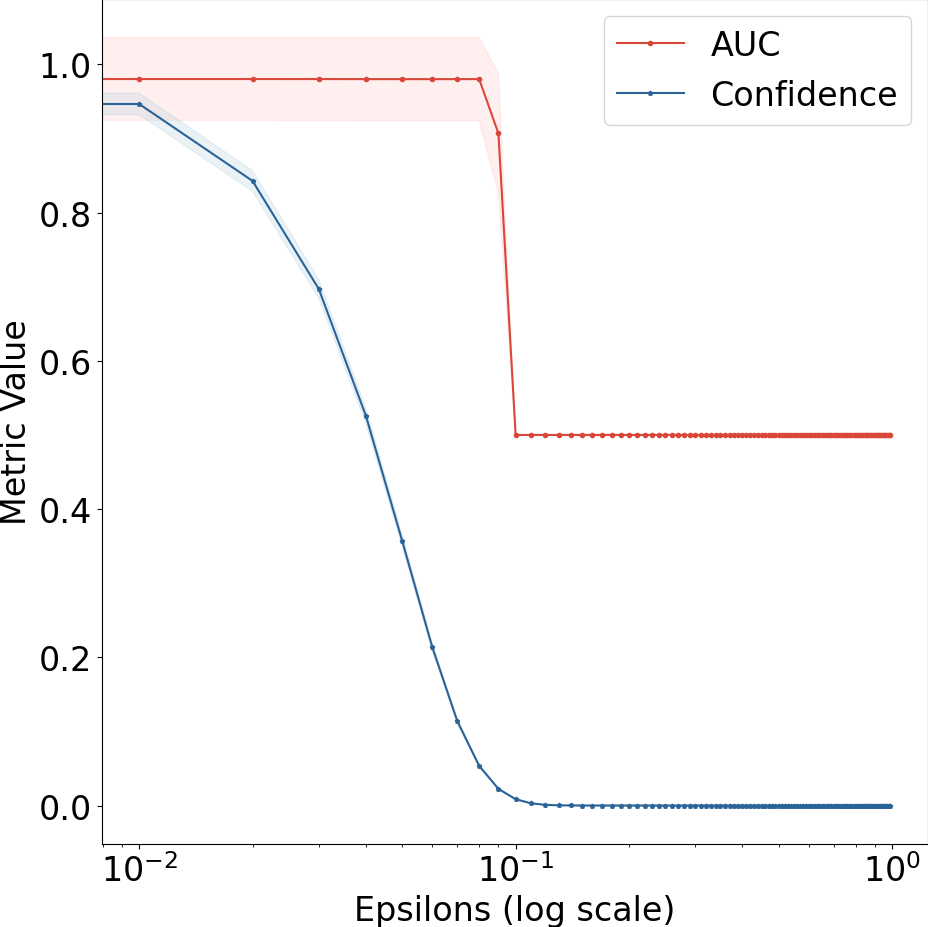}} \\
        { (a)~\mutag} & 
        { (b)~\fluc} & 
        { (c)~\alca} & 
        { (d)~\ben} & 
        { (e)~\bamo} \\
    \end{tabular}
    }
    \caption{Visualization of AUC, NLL, BR, ECE, and confidence score of baseline PGEpxlainer and \ours on five datasets. Each metric is computed at the graph level and plotted with mean and std in the sub-figures. 
    %\hua{better to smooth out the fluctuations in the figures - something like this \url{https://matplotlib.org/stable/_images/sphx_glr_fill_between_alpha_002.png}} 
    The x-axis in the sub-figures denotes the noise level $\epsilon \in [0, 1]$ in the log scale, and the y-axis denotes the value of AUC, NLL, BR, ECE, and confidence score.
    % \jx{@Xiaoou Please tune the font size of the sub-figures, the line width, and the text is unclear to read. Also, use BR to make the terms consistent.}
    }
    \label{fig:corr}
\end{figure*}

\subsection{Experiment Settings}
\subsubsection{Datasets} 
%We focus on ... Thus, we select ... datasets with ground truth explanations in our empirical studies

We evaluate our proposed \ours on a synthetic graph dataset and multiple real-world graph datasets that come with ground-truth explanations
%~\footnote{All the datasets and codes can be found in \url{https://anonymous.4open.science/r/ConfExplainer-E55E}}
. The synthetic dataset, \bamo, is used for graph classification tasks, providing a controlled environment to evaluate explanation methods. The real-world datasets, such as \mutag, \ben, \fluc, and
\alca, consist of molecular graphs and are designed to determine whether a molecule graph contains a specific pattern that affects molecular properties~\cite{agarwal2023evaluating}. These datasets are widely used in GNN explainability and provide a reliable benchmark for assessing the quality of generated explanations.

\label{sec:datasets}

% We introduce G-XAI Bench, a comprehensive data repository and open-source library of Explainable Artificial Intelligence (XAI)-ready datasets designed for evaluating explainers tailored to Graph Neural Networks (GNNs). G-XAI Bench comprises a collection of eleven datasets, including both synthetic and real-world datasets. Notably, several datasets provide ground-truth explanations for assessing GNN explainers. The repository, along with tutorials and reproduction code, is publicly available on GitHub: \url{https://github.com/mims-harvard/GXAI-Bench}.

%\jx{@Xiaoou please introduce the datasets here.}

\noindent~$\bullet$  \textbf{\bamo} A synthetic graph dataset based on the Barabási-Albert (BA) model~\cite{albert2002statistical}, designed for node classification tasks. Each graph contains one of two possible motifs: a 5-node cycle or a 6-node ‘house’ structure, which is randomly attached to a node in a BA base graph. The classification task is to determine whether a given node belongs to one of these motif structures.

\noindent~$\bullet$  \textbf{\mutag}~\cite{kazius2005derivation} A real-world chemistry dataset consisting of molecular graphs for graph-level classification, where nodes represent atoms and edges represent chemical bonds. The objective is to classify molecules on the basis of their mutagenic properties. The presence of specific chemical substructures serves as ground-truth explanations for model predictions.

\noindent~$\bullet$  \textbf{\ben}~\cite{sanchez2020evaluating} A molecular dataset where the task is to classify molecules based on the presence of benzene rings(\chemfig[scale=0.2]{*6(-=-=-=)}).
% Benzene rings are fundamental structural components in many chemical compounds and play a key role in determining molecular properties. 
The ground-truth explanations highlight the benzene substructures responsible for the graph-level classification.

\noindent~$\bullet$  \textbf{\fluc}~\cite{sanchez2020evaluating} A molecular dataset where the objective is to classify molecules based on the presence of fluoride and carbonyl functional groups.
%These functional groups significantly influence the reactivity and behavior of molecules, making them critical factors in the classification process.
The ground-truth explanations focus on identifying these functional groups within the molecular graphs.

\noindent~$\bullet$  \textbf{\alca}~\cite{sanchez2020evaluating} A molecular dataset where the graph-level classification task is based on the presence of an unbranched alkane and a carbonyl functional groups ({C=O}). 
%These chemical groups are essential in many biochemical reactions and influence molecular solubility and acidity. 
The explanations aim to highlight the structure responsible for the classification.

For detailed information on dataset loading and usage, refer to the G-XAI Bench repository~\cite{DVN/KULOS8_2022} and associated documentation.

\subsubsection{Baselines} 
%\jx{@Xiaoou please introduce the baselines here.}

To evaluate the effectiveness of \ours, we compare it against four baseline methods. GNNExplainer (GNNE) and PGExplainer (PGE) are widely used post-hoc explainability methods for GNNs. Additionally, we include two ensemble-based baselines, which are popular techniques for uncertainty estimation. These methods allow us to evaluate both the quality of our explanations and the reliability of our confidence scores.

\noindent~$\bullet$  GNNE~\cite{ying2019gnnexplainer}: A widely used post-hoc explainability method that identifies key sub-graphs responsible for a GNN's predictions. It learns an edge mask by optimizing for fidelity while preserving the model’s original predictions. 
    
\noindent~$\bullet$  PGE~\cite{luo2020parameterized}: A parameterized graph explainer that learns a probabilistic mask distribution over graph edges. Unlike GNNExplainer, PGExplainer incorporates a global view of the dataset by training an explainer network that generalizes across instances.
    
\noindent~$\bullet$  Deep Ensemble~\cite{lakshminarayanan2017deepensembles}: A well-established uncertainty quantification method that trains multiple independent GNNs with different initializations. The ensemble’s variance provides an estimate of prediction uncertainty, but it does not inherently produce explanations for model decisions. We use it as a baseline to compare our confidence-aware explanation quality.

\noindent~$\bullet$  Bootstrap Ensemble~\cite{abdar2021uqreview}: In this approach, multiple models are trained on different bootstrap samples of the dataset. Bootstrapping is particularly useful when the base model lacks intrinsic randomness, as it helps introduce diversity among ensemble members. In our setup, we apply a k-fold ensemble strategy to evaluate the baseline explainers. Specifically, we split the dataset into $k$ fold, train the baseline explainer with $k-1$ fold $k$ times, and evaluate it on the rest $1$ fold. Then we calculate metrics with the mean value of each edge weight on the test graphs with the k results.

    % \item 1. Randomization-based approaches such as random forests [8], where the ensemble members can be trained in parallel without any interaction, and boosting-based approaches where the ensemble members are fit sequentially. 
    
    % \item 2. One of the popular strategies is bagging (a.k.a. bootstrapping), where ensemble members are trained on different bootstrap samples of the original training set. If the base learner lacks intrinsic randomization (e.g. it can be trained efficiently by solving a convex optimization problem), bagging is a good mechanism for inducing diversity. 

    %         In this part, we split the dataset into $k + 1$ fold, train the baseline explainer with $k-1$ fold with $k$ times, and evaluate it on the rest $1$ fold. Then we calculate AUC, NLL, etc. with the mean value of each edge on the test graphs and the confidence interval with the k-results.
    % \item 3. 

\subsubsection{Configurations} 
The experiment configurations are set following prior research~{\cite{zhang2023mixupexplainer}}. A three-layer GCN model was trained
on 80\% of each dataset’s instances as the target model. We apply the same hyper-parameters for all methods: The learning rate was initialized to 0.005, with 30 training epochs. The regularization coefficients are 0.0003 and 0.3. Specifically, the weight of confidence loss for our approach is 100. Explanations are tested in all instances. The experiments are conducted on a Linux server with Ubuntu 16.04, 32 cores Intel(R) Xeon(R) CPU E5-2620 v4 @ 2.10GHz, and NVIDIA TITAN Xp 12 GB GPU.

\subsubsection{Evaluation Metrics} 
%\jx{@Xiaoou please introduce the Evaluation Metrics here.}

To assess the performance of our proposed \ours and baselines, we evaluate the quality of explanations with AUC-ROC score and the reliability of confidence scores using the following metrics: Negative Log-Likelihood (NLL), Briers Score (BR), and Expected Calibration Error (ECE). Detailed description is attached in the Appendix~{\ref{sec:metric}.

\input{tabletexs/table_rq2}

\subsection{Quantitative Evaluation on Explanation Faithfulness (RQ1)}
% In this section, we evaluate the performance and confidence metrics of our proposed approach and other baselines on various datasets. 

% To answer RQ1, we compare our approach with other baseline methods in terms of the AUC-ROC score and confidence metrics used in the literature. The AUC score is evaluated using the weighted vector of the graph generated by the explainers, which serves as the explanation and is compared against the ground truth to calculate the AUC-ROC score. The Confidence metrics, including NLL, BR score, and ECE are also computed with the weighted vector and ground truth. Each experiment is conducted 10 times with random seeds. We summarize the mean performances and std in Table~{\ref{tab:rq1}}.

To address RQ1, we compare our proposed approach with multiple baseline methods, evaluating explanation faithfulness using the AUC-ROC metric. The AUC-ROC score is computed by evaluating the weighted vector of the graph explanation against the ground truth. Each experiment is conducted 10 times with different random seeds, and we report the mean performance and standard deviations in Table~{\ref{tab:auc}}. We have the following observations:
% As shown in Table 1, across all five datasets, with GNNExplainer or PGExplainer as the backbone methods, where we also enhance the confidence of PGExplainer with deep Ensemble and Bootstrap Ensemble, our approach can consistently retrieve the comparable quality of obtained explanations and shows a good confidence value.  Specifically, our improves the AUC scores by xxx, on average, on the xxx datasets, and xxx on xxx. And the confidence metrics show that xxx.

\noindent~$\bullet$ In \bamo, \mutag, \fluc, and \alca, our method achieves the best AUC-ROC scores compared to baseline explainability methods, including GNNExplainer, PGExplainer, and ensemble approaches. Our model significantly outperforms these baselines in explanation quality, demonstrating its ability to produce more faithful and interpretable graph explanations. 

\noindent~$\bullet$ Specifically, our approach improves AUC scores by 21.00\% and 24.52\% on average, with particularly strong performance gains on datasets such as \bamo and \mutag. This highlights that our confidence-aware framework generates more accurate and reliable explanations, aligning better with the underlying ground truth. 

\noindent~$\bullet$ In \ben dataset, the Deep Ensemble approach retrieved the best AUC, but our method still has a comparable performance, since our method emphasizes confidence scoring while maintaining comparable accuracy.

\subsection{Confidence Scoring and Calibration}

\begin{figure*}[]
    \centering
     \includegraphics[width=1\textwidth]{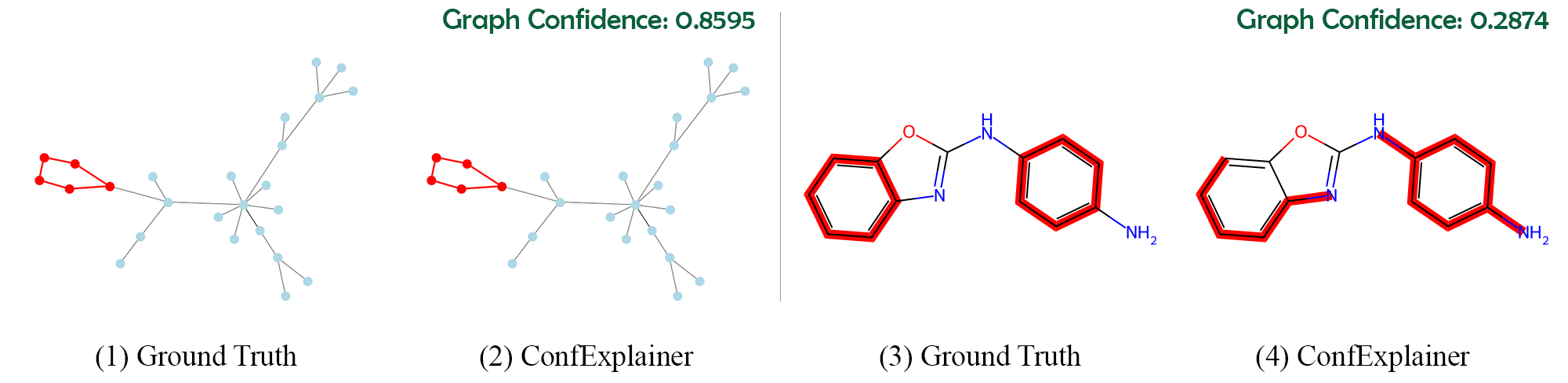}
     % \vspace{-0.1cm}
     \caption{Visualization of two graph case studies using \ours. On the left (1 and 2), with samples from the \bamo dataset, our model successfully identifies the "house" structure, highlighted in red, with a relatively high confidence score (0.8595). On the right (3 and 4), with samples from the Benzene dataset, our model fails to accurately detect the significant substructure in the chemical graph with a lower confidence score (0.2874).}
     \label{fig:case_study}
     % \vspace{-0.1cm}
\end{figure*}

In this section, we answer the RQ2 with two parts of experiments: (1) we compared our method with baseline methods on the confidence metrics including NLL, BR, and ECE. This a static experiment, where we generate the explanation for original graph instances and compute the metrics with ground truth labels. The results are shown in Table~{\ref{tab:uncertainty}}, more analysis is conducted below;
% \hua{from table 2 seems like our method is not good, need more explanation why} 
(2) to further demonstrate the effectiveness of our proposed method, we introduce the OOD scenario by injecting the noise into the original graphs. With the increasing noise level, we visualize the explanation accuracy and model confidence. Specifically, we add noise in the edge feature level by disturbing the node feature with a consistent value $\epsilon$. The higher $\epsilon$ is, the more OOD the original graph is, which would make it harder for trained models to recognize the explanation sub-graph, and the explainer confidence should decrease as well. 
The results are shown in Figure~\ref{fig:corr}, wherein the first row, we plot the AUC score and confidence evaluation metrics, including NLL, BR, and ECE of plain PGExplainer; in the second row, we visualize our proposed approach with AUC and confidence score generated by the confidence scoring model.

\begin{figure*}[ht]
    \centering
     \includegraphics[width=1\textwidth]{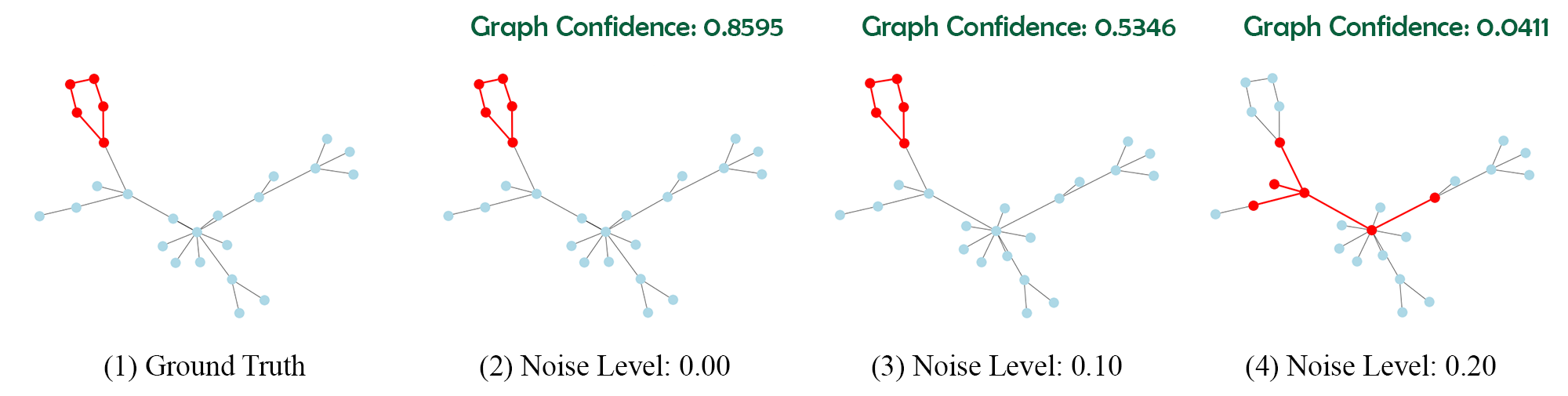}
     % \vspace{-0.1cm}
     \caption{Visualization of the impact of different noise levels on the confidence scores and explanations provided by \ours. This figure shows that as the noise level increases from 0.00 to 0.20, the model's confidence score drops significantly from 0.8595 to 0.5346, and finally to 0.0411. At higher noise levels, the model faces difficulty in accurately identifying the ground truth structure, as illustrated in (4).}
     \label{fig:Noise Attack}
     % \vspace{-0.1cm}
\end{figure*}

We have the following observations: 

\noindent~$\bullet$ The confidence score could better evaluate the confidence of the explainer model, as it shows a decreasing tendency while adding more noise and the performance of the explainer is dropping, as shown in Figure~\ref{fig:corr}, while NLL and BR couldn't reflect. Moreover, the confidence score decreases smoothly when more noise is injected, while ECE only drop-down at the critical point; 
% \noindent~$\bullet$ Other confidence evaluation metrics all require the ground truth explanation to compute the score, while the confidence scoring model doesn't need it, which means the confidence model could help evaluate the confidence of the explainer on the OOD or new coming dataset without ground truth explanation. 
The reason for \ours not performing best in Table~{\ref{tab:uncertainty}} is these metrics rely more on the concept of uncertainty, which is different from confidence, and it's not specifically optimized by our approach.

\noindent~$\bullet$ Unlike previous confidence evaluation metrics, which require access to ground truth explanations for score computation, our proposed confidence scoring model operates independently of ground truth. This key advantage enables our approach to effectively assess the confidence of explanations in out-of-distribution (OOD) scenarios or on newly encountered datasets, where ground truth explanations may be unavailable. 
The reason our method does not achieve the highest performance in Table~{\ref{tab:uncertainty}} is that existing uncertainty-based metrics are primarily designed to capture prediction uncertainty, which is conceptually distinct from explanation confidence. Since our approach is not explicitly optimized for uncertainty estimation, these metrics do not fully reflect the improvements brought by our confidence-aware framework. 
% Instead, our method is specifically designed to provide direct confidence assessments for GNN explanations, making it more suitable for real-world deployments where ground truth labels are unavailable.

\noindent~$\bullet$ By training the explainer with the confidence model, the explainer is more robust on the slight noise disturbing. The rash point is larger for \ours than the baseline PGExplainer, and the converged AUC value is 0.5, a random result, compared to the 0.0 in the first row, where PGExplainer gives the fully opposite results.
In conclusion, our confidence model effectively evaluates the explainer's confidence and improves its robustness.

\subsection{Case Study: Could the confidence score reveal the explainer's performance? (RQ3)}

In this section, we showcase the confidence score and explanation from the explainer (\ours) in two perspectives: The first part includes the static result, where we visualize the result from different graphs; the second part includes the noise injection, where we visualize the compression of one graph with different levels of the noise. Both part is drawn with the original graph, highlighted top-k explanation edges, and paired confidence score.

\subsubsection{Static Study} 
As shown in Figure~\ref{fig:case_study}, the left and right sides visualize examples from two different datasets. Figure~\ref{fig:case_study}(1) and Figure~\ref{fig:case_study}(2) correspond to the \bamo dataset, where the explanation is both accurate and associated with a high confidence score. In contrast, Figure~\ref{fig:case_study}(3) and Figure~\ref{fig:case_study}(4) correspond to the \ben dataset, where the explanation is discrete and accompanied by a low confidence score. In Figure~\ref{fig:case_study}(1) and Figure~\ref{fig:case_study}(3), the red-highlighted edges represent the ground-truth explanations. Figure~\ref{fig:case_study}(2) and Figure~\ref{fig:case_study}(4) illustrate the results produced by \ours, including the predicted explanations (highlighted) and the corresponding graph confidence scores. This observation aligns with the results presented in the previous table, demonstrating that the confidence module effectively evaluates the reliability of model explanations. Additional examples from more datasets can be found in the appendix Figure~\ref{fig:appendix_casestudy}.

%case (a) has an accurate explanation as well as high confidence, while case (b) has a discrete explanation together with a low confidence score. This observation support the results in the previous table and demonstrates the ability of the confidence module to evaluate the model performance.

% \begin{figure}[h!]
%     \centering
%     % First subfigure
%     \subfigure[Case a]{
%         \includegraphics[width=0.2\textwidth]{figures/case_study/cs1_1.png}
%         \label{fig:cs1_1}
%     }
%     % Second subfigure
%     \subfigure[Case b]{
%         \includegraphics[width=0.4\textwidth]{figures/case_study/cs1_2.png}
%         \label{fig:cs1_2}
%     }
%     \caption{Visualization of Correlations.}
%     \label{fig:cs1}
% \end{figure}

% \begin{figure}[h!]
%     \centering
%     \subfigure[Case a]{
%         \includegraphics[width=0.2\textwidth]{figures/case_study/ba2motif_1.png}
%         \label{fig:cs1_1}
%     }
%     \hspace{0.05\textwidth}
%     \subfigure[Case b]{
%         \includegraphics[width=0.2\textwidth]{figures/case_study/ba2motif_2.png}
%         \label{fig:cs1_2}
%     }
%     \caption{Visualization of two graph case studies in ba-2motifs dataset. In Case a, our model successfully detects the "house" structure, as highlighted in red, with a relative high confidence score. In Case b, our model fails to accurately identify the "house" structure, with a lower confidence score.}
%     \label{fig:cs1}
% \end{figure}

\subsubsection{Noise Injection} 
To visualize the effect of the noise injection, we conduct an experiment on injecting the noise into the graph embedding before feeding it into the explainer, and observing the change of the explanation fidelity and confidence. We randomly sample a graph from the dataset \bamo for the case study. As observed in Figure~\ref{fig:Noise Attack}. As more noise is injected into the graph embedding, the explainer’s confidence decreases steadily, while the explanation accuracy initially remains stable but eventually collapses at a higher noise level, when the noise level increase from 0.1 to 0.2, aligning with the observation in Figure~{\ref{fig:corr}}. It's aligned with the hypothesis that with the noise injection, the explainers' performance and confidence would decrease in parallel, indicating the effectiveness of our confidence evaluation module.

%% file: tabletexs/table1_rq1.tex
\begin{table*}[ht]
\centering
\caption{Comparison of AUC-ROC ($\uparrow$) performance on five datasets. The best results are highlighted in bold. The second-best results are underlined. \ours achieves the best performance in \bamo, \mutag, \fluc and \alca, while holds a comparable performance in \ben. 
%\hua{might also highlight second best with an underline. Need to put observations in the caption as well to make it self-contained.} 
}
\label{tab:auc}
\begin{scalebox}{1.0}{
    \begin{tabular}{cccccc}
        \hline
         Dataset  &  GNNE &  PGE  &  Deep Ensemble  &  Bootstrap Ensemble  &  \ours \\
        \hline
        \multirow{1}{*}{\bamo} 
                     & 54.58\% $\pm$ 1.21\% & \underline{88.15\% $\pm$ 0.69\%} & 80.47\% $\pm$ 1.20\% & 81.56\% $\pm$ 0.96\% & \textbf{97.19\% $\pm$ 0.04\%} \\
        \multirow{1}{*}{\mutag} 
                     & 59.77\% $\pm$ 0.67\% & 65.31\% $\pm$ 8.78\% & \underline{68.78\% $\pm$ 1.16\%} & 68.63\% $\pm$ 1.53\% & \textbf{90.14\% $\pm$ 0.97\%} \\
        \multirow{1}{*}{\ben} 
                     & 50.40\% $\pm$ 0.30\% & 77.57\% $\pm$ 1.43\% & \textbf{78.60\% $\pm$ 0.24\%} & \underline{78.11\% $\pm$ 0.24\%} & 77.80\% $\pm$ 1.32\% \\
        \multirow{1}{*}{\fluc} 
                     & 54.72\% $\pm$ 0.67\% & 54.78\% $\pm$ 4.42\% & 52.59\% $\pm$ 1.11\% & \underline{55.70\% $\pm$ 1.61\%} & \textbf{59.27\% $\pm$ 1.48\%} \\
        \multirow{1}{*}{\alca} 
                     & 54.03\% $\pm$ 2.13\% & 73.05\% $\pm$ 2.00\% & 74.44\% $\pm$ 0.18\% & \underline{75.17\% $\pm$ 0.33\%} & \textbf{81.52\% $\pm$ 0.22\%} \\
        \hline
    \end{tabular}
}
\end{scalebox}
\end{table*}

%% file: tabletexs/table_rq2.tex
\begin{table*}[h]
\centering
\caption{Comparison of NLL ($\downarrow$), BR ($\downarrow$), and ECE ($\downarrow$) on five datasets. Lower values indicate better performance. % \hua{Avoid putting this table along with Table 1, which might confuse the reviewers about the ``main table'', might also highlight second best with an underline. Need to put observations in the caption as well to make it self-contained. We might also need to say the conclusion as well } 
}
\label{tab:uncertainty}
\begin{scalebox}{1.0}{
    \begin{tabular}{ccccccc}
        \hline
         Dataset  & Metric  &  GNNE &  PGE  &  Deep Ensemble  &  Bootstrap Ensemble  &  \ours \\
        \hline
        \multirow{3}{*}{\bamo} 
                    & NLL & 0.7197 $\pm$ 0.0005 & 0.6318 $\pm$ 0.0081 & \underline{0.6171 $\pm$ 0.0007} & 0.6193 $\pm$ 0.0012 & \textbf{0.6117 $\pm$ 0.0008} \\
                    & BR  & \textbf{0.2449 $\pm$ 0.0011} & \underline{0.5705 $\pm$ 0.0397} & 0.7205 $\pm$ 0.0046 & 0.6996 $\pm$ 0.0085 & 0.7340 $\pm$ 0.0177 \\
                    & ECE & \textbf{0.2714 $\pm$ 0.0009} & \underline{0.6553 $\pm$ 0.0290} & 0.7480 $\pm$ 0.0023 & 0.7337 $\pm$ 0.0062 & 0.7858 $\pm$ 0.0100 \\
        \hline
        \multirow{3}{*}{\mutag} 
                    & NLL & 0.6965 $\pm$ 0.0000 & 0.6832 $\pm$ 0.0004 & \underline{0.6831 $\pm$ 0.0000} & 0.6832 $\pm$ 0.0000 & \textbf{0.6829 $\pm$ 0.0001} \\
                    & BR  & \textbf{0.2413 $\pm$ 0.0001} & \underline{0.9275 $\pm$ 0.0276} & 0.9335 $\pm$ 0.0022 & 0.9282 $\pm$ 0.0017 & 0.9375 $\pm$ 0.0125 \\
                    & ECE & \textbf{0.4630 $\pm$ 0.0001} & \underline{0.9490 $\pm$ 0.0149} & 0.9523 $\pm$ 0.0012 & 0.9496 $\pm$ 0.0009 & 0.9546 $\pm$ 0.0065 \\
        \hline
        \multirow{3}{*}{\ben} 
                    & NLL & 0.7143 $\pm$ 0.0001 & 0.6203 $\pm$ 0.0004 & \underline{0.6202 $\pm$ 0.0001} & 0.6203 $\pm$ 0.0001 & \textbf{0.6201 $\pm$ 0.0004} \\
                    & BR  & \textbf{0.2620 $\pm$ 0.0003} & 0.7931 $\pm$ 0.0063 & \underline{0.7887 $\pm$ 0.0012} & 0.7910 $\pm$ 0.0010 & 0.7910 $\pm$ 0.0061 \\
                    & ECE & \textbf{0.3158 $\pm$ 0.0003} & 0.7985 $\pm$ 0.0036 & \underline{0.7960 $\pm$ 0.0007} & 0.7973 $\pm$ 0.0006 & 0.7973 $\pm$ 0.0035 \\
        \hline
        \multirow{3}{*}{\fluc} 
                    & NLL & 0.6979 $\pm$ 0.0001 & 0.6789 $\pm$ 0.0001 & \textbf{0.6789 $\pm$ 0.0000} & \underline{0.6789 $\pm$ 0.0001} & 0.7166 $\pm$ 0.0001 \\
                    & BR  & \underline{0.2393 $\pm$ 0.0005} & 0.9471 $\pm$ 0.0075 & 0.9460 $\pm$ 0.0011 & 0.9442 $\pm$ 0.0010 & \textbf{0.0378 $\pm$ 0.0002} \\
                    & ECE & \underline{0.4446 $\pm$ 0.0005} & 0.9541 $\pm$ 0.0039 & 0.9535 $\pm$ 0.0006 & 0.9526 $\pm$ 0.0006 & \textbf{0.0318 $\pm$ 0.0025} \\
        \hline
        \multirow{3}{*}{\alca} 
                    & NLL & 0.6959 $\pm$ 0.0001 & \textbf{0.6884 $\pm$ 0.0007} & \underline{0.6887 $\pm$ 0.0001} & 0.6888 $\pm$ 0.0001 & 0.7052 $\pm$ 0.0002 \\
                    & BR  & \underline{0.2422 $\pm$ 0.0015} & 0.5051 $\pm$ 0.0822 & 0.4623 $\pm$ 0.0088 & 0.4662 $\pm$ 0.0100 & \textbf{0.0213 $\pm$ 0.0002} \\
                    & ECE & \underline{0.4643 $\pm$ 0.0015} & 0.6762 $\pm$ 0.0635 & 0.4652 $\pm$ 0.0080 & 0.6502 $\pm$ 0.0090 & \textbf{0.0094 $\pm$ 0.0018} \\
        \hline
    \end{tabular}
}
\end{scalebox}
% \vspace{-0.2cm}
\end{table*}

%% file: conclusion.tex
\section{Conclusion}

In this work, we addressed a critical gap in Graph Neural Network (GNN) explainability by introducing a confidence-aware explanation framework, Generalized Graph Information Bottleneck with Confidence Constraints (GIB-CC). While existing GNN explainers provide insights into model decisions, they often fail to quantify the trustworthiness of their explanations. Our approach explicitly integrates confidence estimation into the explanation process, ensuring that users can assess not only what a model explains but also how reliable that explanation is.
Through extensive theoretical analysis and empirical validation, we demonstrated that our framework significantly improves both explanation faithfulness and confidence calibration. Our confidence evaluation module enhances explanation reliability, enabling better differentiation between high-fidelity and low-fidelity explanations, particularly in out-of-distribution (OOD) scenarios. Moreover, our confidence-aware training strategy improves the robustness of GNN explainers, mitigating performance degradation under distribution shifts.

Looking forward, our work paves the way for further trustworthy AI research in graph-based learning. Future directions include extending our confidence-aware framework to dynamic and evolving graphs, exploring more efficient confidence estimation techniques, and integrating confidence-aware explanations into real-world applications such as drug discovery, fraud detection, and decision support systems. By bridging the gap between explainability and reliability, we hope to contribute to the broader adoption of trustworthy and interpretable GNNs in high-stakes domains.

%% file: appendix.tex
\newpage
\clearpage
\appendix
\section{Notation Table}

\begin{table}[h]
    \centering
    \begin{tabular}{c l}
        \hline
        \textbf{Symbol} & \textbf{Description} \\
        \hline
        $G= (\mathcal{V}, \mathcal{E}; \mX, \mA)$ & Original Graph \\
        $\mathcal{E}$ & Edge set\\
        $\mX$ & Feature matrix\\
        $\mA$ & Adjacency matrix\\
        $\mM$ & Binary mask \\
        $G'$ & Candidate sub-graph \\
        $\tilde G$ & Calibrated sub-graph \\
        $ G^r $ & Gaussian noise\\
        $Y$ & Target label \\
        $\tilde Y$ & Predicted label \\
        $A$ & Adjacency matrix of the graph \\
        $I(\cdot)$ & The mutual information \\
        $G^*$ & Explanation sub-graph \\
 
        $\mC$ & Confidence vector \\
        $c_i$ & Confidence score on each edge\\
        $x_i $ & Raw confidence score from MLP \\
        $\lambda$ & Balance between explanation and confidence \\
        $\odot$ & Element-wise product \\
        $H(\cdot)$ & Entropy \\
        $CE(\cdot)$ & Cross-Entropy loss\\
        $\mathcal{L}$ & Loss function \\
        $f(\cdot)$ & Confidence model \\
        $ e $ & Hidden state embedding \\
        $\gD$ & Dimension \\
        $\gO$ & Time complexity\\
        %$z^*$ & Confidence level \\
        $\mathcal{B}_k$ & Bin \\ 
        $[l_k, u_k]$ & Prediction confidence interval \\
        $C_{AN}$ & Correlation between AUC and NLL \\
        $C_{AB}$ & Correlation between AUC and BR \\
        $C_{AE}$ & Correlation between AUC and ECE \\
        \hline
    \end{tabular}
    \caption{Notation Table}
\end{table}

\section{Justification of $I(C, G^r; Y \mid G^{\prime}) = 0$}
\label{sec:justification_core_condition}
The core assumption $I(C, G^r; Y \mid G^{\prime}) = 0$ underpins the theoretical equivalence between GIB and GIB-CC. To justify this condition, we begin with the fact that $G^r$ is sampled as independent random Gaussian noise, and hence shares no mutual information with $Y$ or $G^{\prime}$. This leads to the simplification:
$$I(C, G^r; Y \mid G^{\prime}) = I(C; Y \mid G^{\prime})$$
We further expand this mutual information term:
$$I(C; Y \mid G^{\prime}) = H(C \mid G^{\prime}) - H(C \mid Y, G^{\prime})$$
The confidence matrix C is defined as:
$$C = \text{MLP}(f_{\text{emb}}(G), G^{\prime}),$$
where $f_{\text{emb}}(G)$ is the embedding of the input graph. This implies that any dependence between $C$ and $Y$ must pass through $G$. Under the vanilla GIB assumption, we have
$$I(Y; G \mid G^{\prime}) = 0 \quad i.e., G^{\prime}$$ 
capturing all label-relevant information from $G$. Since $C$ is a deterministic function of $G$ and $G^{\prime}$, it follows that once $G^{\prime}$ is given, $C$ cannot capture additional information about $Y$, yielding:
$I(C; Y \mid G^{\prime}) = 0$
Thus, our assumption $I(C, G^r; Y \mid G^{\prime}) = 0$ is a natural extension of the GIB formulation under its own information bottleneck condition.

Moreover, the condition $I(C, G^r; Y \mid G^{\prime}) = 0$ may not strictly hold in all cases — for example, under severe OOD shifts where $G^{\prime}$ fails to retain label-relevant information. However, this limitation originates from the standard GIB formulation itself, rather than our proposed extension.  Weaker or relaxed variants of this assumption would be explored in future work to enhance robustness in more general settings.

\section{Training Procedure} 
\label{sec:algo}
Since the confidence loss contains both confidence score and explanation mask, to avoid of trivial solution, we train the explainer model and confidence evaluator model iteratively. When training one of them, another one is frozen. Our training algorithm is shown in Algorithm~\ref{alg:train}.

\renewcommand{\algorithmicrequire}{\textbf{Input:}}
\renewcommand{\algorithmicensure}{\textbf{Output:}}

\begin{algorithm}
	\caption{Training Algorithm} 
	\begin{algorithmic}[1]
        \Require Target to-be-explained graph set $\mathcal{G}$, to-be-explained GNN model $f$, explainer model $E$, confidence model $c$, training epochs $e$.
       \Ensure Trained explainer $E$ and confidence evaluator $c$.
        \State Initialize explainer model $E$ and confidence evaluator model $c$.
        \For {$epoch \in 2*e$}
            \For{$G \in \mathcal{G}$}
                \State $G^r \gets $ Randomly sampled graph Gaussian noise from $\mathcal{N}(0, 1)$
                \State $H \gets f(G)$
                \State $G^* \gets E(H)$
                \State $C \gets c(H, G^*)$
                \State $\tilde G \gets CG^* + (1-C) G^r$
                \State Compute $\mathcal{L}_{\text{GIB}}(G, G^*, \tilde G)$
                \State Compute $\mathcal{L}_{C}(C, G^*)$
            \EndFor
            \If{$epoch \% 2 == 0$}
            \State Update $E$ with back propagation.
        \Else
            \State Update $c$ with back propagation.
        \EndIf
        \EndFor
        \State \Return Explainer $E$, confidence evaluator $c$
    \end{algorithmic} 
    \label{alg:train}
\end{algorithm}

\section{Evaluation Metrics} 
%\jx{@Xiaoou please introduce the Evaluation Metrics here.}
\label{sec:metric}
To assess the performance of \ours and baselines, we evaluate the quality of explanations with AUC-ROC score and the reliability of confidence scores using the following metrics: Negative Log-Likelihood (NLL), Briers Score(BS), and Expected Calibration Error(ECE):

\begin{itemize}
    \item Area Under the Receiver Operating Characteristic Curve (AUROC): Measures the quality of explanations generated by the explainer. A higher AUROC indicates that the method effectively identifies meaningful substructures contributing to predictions.
    \item Negative Log-Likelihood (NLL): Evaluates the confidence scores by measuring how well the predicted probabilities align with the ground truth. The NLL is defined as: $NLL = - \frac{1}{N}\sum^N_{i=1}\log p_i$
    , where $N$ is the total number of samples and $p_i$ is the predicted probability for each sample. Lower NLL values indicate better confidence estimation performance.
    % \item Confidence Interval: 
    
    % Confidence interval = sample mean ± margin of error. 
    
    % $ME = z^* \cdot \frac{\text{population standard deviation}}{\sqrt{n}} = z^* \cdot \text{std}$, where $z^*$ is the confidence level.

    \item Briers Score(BR)~\cite{brier1950bscore}: BS is the mean squared error between the predicted probability and ground-truth label 
    
    $\frac{1}{K} \sum_y{(p_\theta (y|x_n) - \mathbf{1}_{y=y_n})^2}$, where $\mathbf{1}$ is the indicator function. It serves as another proper scoring rule for classification tasks.Lower values indicating better performance.

    \item Expected Calibration Error(ECE): Measures the discrepancy between predicted confidence and actual accuracy across different confidence bins.  $\text{ECE} = \underset{k=1}{\overset{K}{\sum}}\frac{|\mathcal{B}_k|}{n}\hat{\varepsilon}_k$, 
    
    $\hat{\varepsilon}_k = \frac{1}{\mathcal{B}_k}\left|\underset{i\in \mathcal{B}_k}{\sum}\left[\mathbb{1}(\hat{y}_i == y_i) - p_i\right]\right|$, where $p \in [l_k, u_k]$, $B_k$ denotes the bin with prediction confidences bounded between $l_k$ and $u_k$. A lower ECE suggests better confidence scores.
    
    % \item Relative Confidence: similar to confidence interval

    \item Pearson Correlation: Measures the correlation between confidence scores and explanation quality. A higher correlation indicates that the explainer's confidence aligns well with the actual quality of explanations, ensuring reliability in uncertainty estimation.
    
    %Specifically, for PGExplainer as a baseline, we calculate the correlation between predicted edge probability and the ground truth. We calculate the correlation between the confidence score and AUC separately. 

\end{itemize}

\section{Time Complexity}
In addition to the inference cost, this section contains a complete time complexity analysis and empirical evaluation of the training overhead of the full framework.

Formally, the training complexity consists of: 1. The explainer and confidence evaluator, both implemented as MLPs operating on graph embeddings and masks, each with complexity $\mathcal{O}(|E|d)$, where $|E|$ is the number of edges and d is the embedding dimension. 2. The calibrated graph construction involves a weighted sum: $\tilde{G} = C \odot G^{\prime} + (1-C) \odot G^r$, with complexity $\mathcal{O}(|E|)$. 3. Total training cost per iteration is thus linear in the graph size: $\mathcal{O}(|E|d)$, consistent with prior works like PGExplainer.

\begin{table*}[h]
    \centering
    \begin{tabular}{c | c c c c}
        \hline
        \textbf{Dataset} & \textbf{PGE} & \textbf{PGEBE} & \textbf{PGEDE} & \textbf{\ours} \\
        \hline
        \bamo  & 283.11      & 1081.05     & 1342.88     & 534.23       \\
        \mutag     & 2598.70     & 8532.29     & 12122.73    & 4239.88     \\
        \ben   & 3274.48     & 12425.79    & 15630.53    & 7043.03     \\
        \fluc      & 2394.31     & 9056.82     & 11324.38    & 5034.04     \\
        \alca      & 311.45      & 1136.57     & 1407.92     & 637.18      \\
        \hline
    \end{tabular}
    \caption{Training Time Cost}
    \label{tab:training_time_cost}
\end{table*}

Table~\ref{tab:training_time_cost} compares wall-clock training time with PGExplainer and PGExplainer with Bootstrap-Ensemble/Deep-Ensemble in all datasets. As shown in the table, \ours achieves comparable efficiency to PGExplainer and is significantly faster than ensemble-based methods.

% | Dataset   | PGE         | PGEBE       | PGEDE       | \ours    |
% |-----------|-------------|-------------|-------------|-------------|
% | ba2motif  | 283.11      | 1081.05     | 1342.88     | 534.23      |
% | alca      | 311.45      | 1136.57     | 1407.92     | 637.18      |
% | mutag     | 2598.70     | 8532.29     | 12122.73    | 4239.88     |
% | benzene   | 3274.48     | 12425.79    | 15630.53    | 7043.03     |
% | flca      | 2394.31     | 9056.82     | 11324.38    | 5034.04     |

\subsection{Ablation Study}

\begin{figure}[ht]
    \centering
    \includegraphics[width=0.5\textwidth]{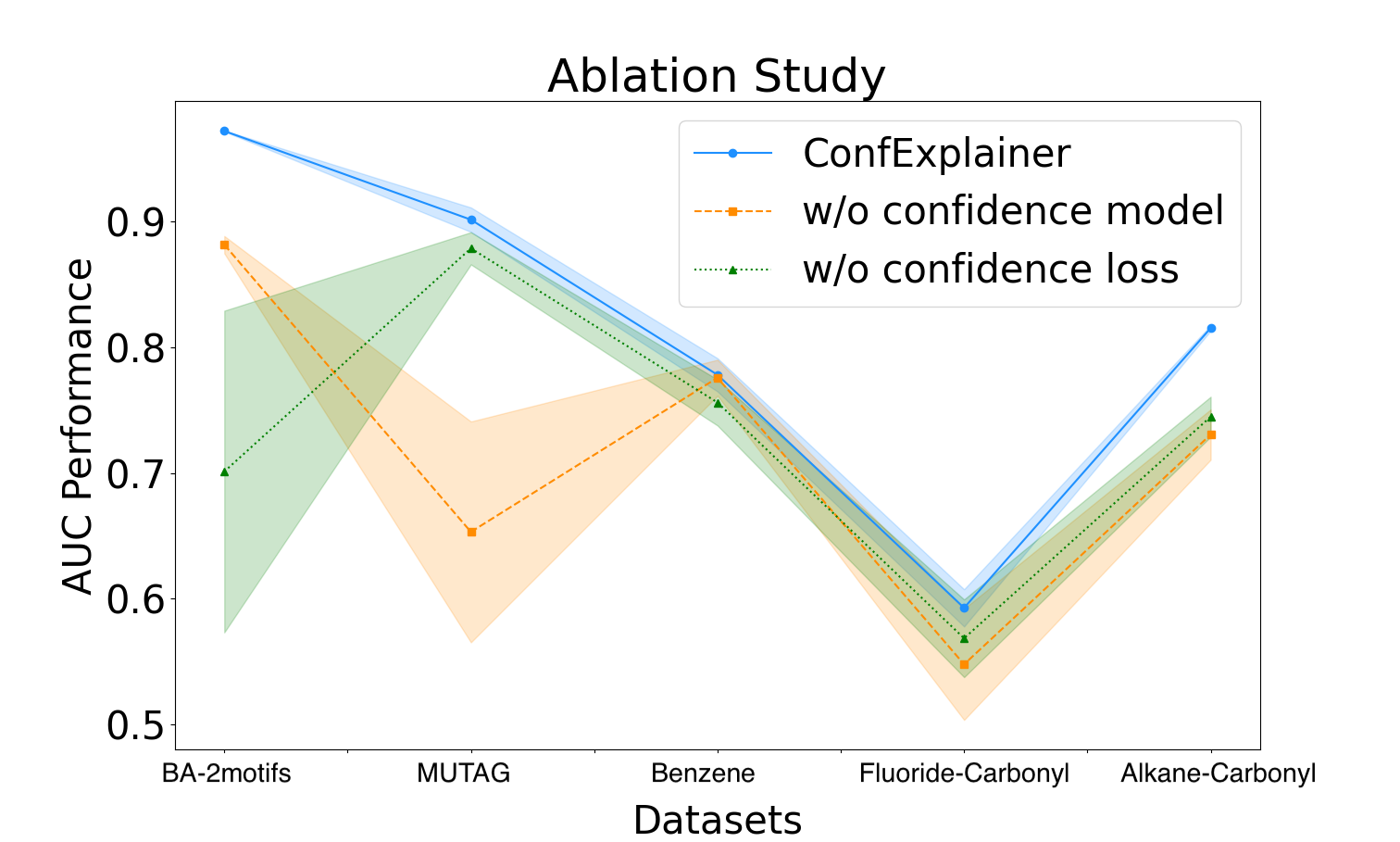}
    \caption{Visualization of ablation study}
    \label{fig:ablation}
    \vspace{-0.6cm}
\end{figure}

% In this section, we conduct the ablation study on our proposed approach. We decomposed our approach into two variants: (1): w/o confidence module: we removed the confidence evaluation module and downgrade our explainer into a general explainer. (2): w/o ECE loss: we removed the ECE loss but keep the confidence module, optimized it together with the GIB. As show in the Figure, the results indicate the effectiveness of the modules design, which could help our approach maintain a comparable performance while leveraging the confidence evaluation.

To assess the contribution of each component in our proposed approach, we perform an ablation study by decomposing the model into two key variants:

\noindent~$\bullet$ w/o Confidence Module: In this variant, we remove the confidence evaluation module, effectively downgrading our model to a general explainer without confidence assessment.

\noindent~$\bullet$ w/o Confidence Loss: Here, we retain the confidence evaluation module but eliminate the confidence loss, optimizing the module jointly with the Graph Information Bottleneck (GIB) framework.

The results, as shown in Figure~{\ref{fig:ablation}}, demonstrate the effectiveness of our module design. Specifically, the confidence module plays a crucial role in enhancing explanation reliability, while the inclusion of confidence loss further refines the confidence calibration. Notably, our approach maintains comparable performance to the baseline explainer while benefiting from the additional confidence evaluation, validating the necessity of both components.

\section{Hyper-parameter Tuning}
To justify the choice of the confidence loss weight $\lambda$, a sensitivity analysis across multiple datasets was conducted by varying $\lambda$ in $[0.001, 0.01, 0.1, 1, 10, 100, 1000]$. The results of AUC, reported in Table~{\ref{tab:Hyper-parameter-Tuning}}, show that \ours is robust across a wide range of $\lambda$ values. We chose $\lambda=100$ as it provides consistently strong performance across datasets.

\begin{table*}[h]
    \centering
    \begin{tabular}{c | c c c c c c c c }
        \hline
        \textbf{Dataset} & 0 & 0.001 & 0.01 & 0.1 & 1 & 10 & 100 & 1000 \\
        \hline
        \bamo  & 0.7011  & 0.8271  & 0.9717  & 0.9717  & 0.9719  & 0.9719  & 0.9719  & 0.9717  \\
        \mutag     & 0.6226  & 0.9012  & 0.9011  & 0.9012  & 0.9014  & 0.9017  & 0.9014  & 0.9011  \\
        \ben   & 0.7560  & 0.7578  & 0.7694  & 0.7691  & 0.7690  & 0.7890  & 0.7780  & 0.7690  \\
        \fluc      & 0.7036  & 0.7691  & 0.5959  & 0.5928  & 0.5927  & 0.5927  & 0.5927  & 0.5927  \\
        \alca      & 0.7768  & 0.8099  & 0.8055  & 0.8059  & 0.8070  & 0.8275  & 0.8152  & 0.8133  \\
        \hline
    \end{tabular}
    \caption{Hyper-parameter Tuning}
    \label{tab:Hyper-parameter-Tuning}
\end{table*}

% | Dataset   | 0       | 0.001   | 0.01    | 0.1     | 1       | 10      | 100     | 1000    |
% |-----------|---------|---------|---------|---------|---------|---------|---------|---------|
% | ba2motif  | 0.7011  | 0.8271  | 0.9717  | 0.9717  | 0.9719  | 0.9719  | 0.9719  | 0.9717  |
% | mutag     | 0.6226  | 0.9012  | 0.9011  | 0.9012  | 0.9014  | 0.9017  | 0.9014  | 0.9011  |
% | benzene   | 0.7560  | 0.7578  | 0.7694  | 0.7691  | 0.7690  | 0.7890  | 0.7780  | 0.7690  |
% | flca      | 0.7036  | 0.7691  | 0.5959  | 0.5928  | 0.5927  | 0.5927  | 0.5927  | 0.5927  |
% | alca      | 0.7768  | 0.8099  | 0.8055  | 0.8059  | 0.8070  | 0.8275  | 0.8152  | 0.8133  |

\section{Figures}

\subsection{Visualization of the confidence and explanations over datasets}

\begin{figure*}[h!]
    \centering
    \subfigure[\ben dataset]{
        \includegraphics[width=1.0\textwidth]{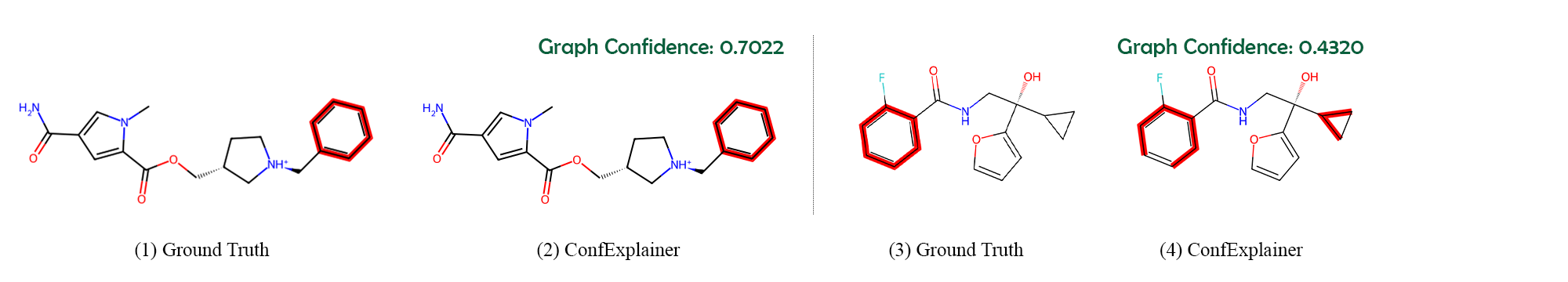}
        \label{fig:benzene_conf}
    }
    \subfigure[\fluc dataset ]{
        \includegraphics[width=1.0\textwidth]{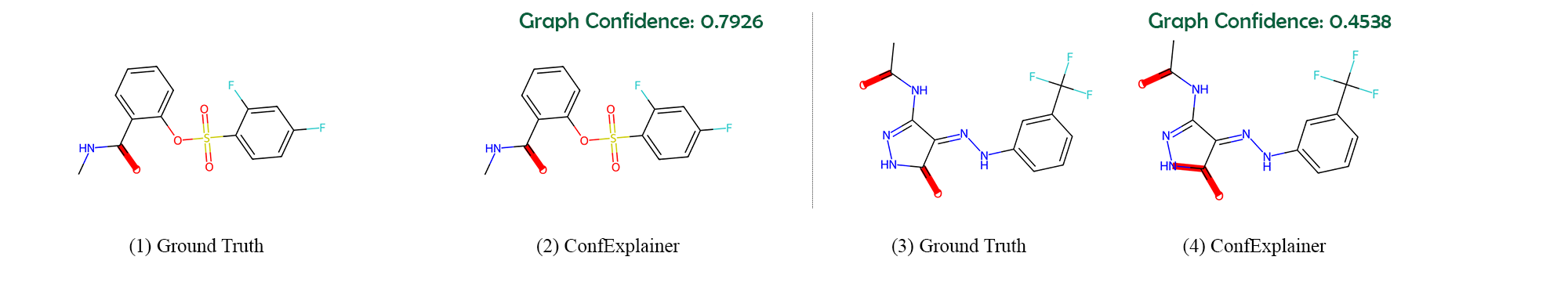}
        \label{fig:flca_conf}
    }
     \subfigure[\alca dataset]{
        \includegraphics[width=1.0\textwidth]{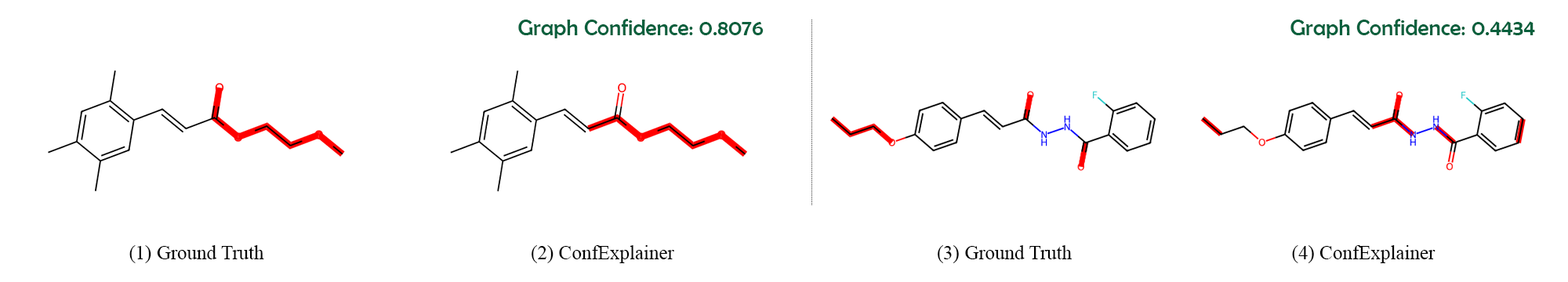}
        \label{fig:alca_conf}
    }
    \caption{Visualization of additional examples from the \ben, \fluc, and \alca datasets, with the highlighted regions indicating the explanations and graph confidence scores.}
    \label{fig:appendix_casestudy}
\end{figure*}

This section provides additional visualizations of ConfExplainer on different datasets. Figure~(\ref{fig:appendix_casestudy}) includes examples from three datasets:(a) Benzene dataset, (b) Fluoride-Carbonyl dataset, and (c) Alkane-Carbonyl dataset.

Each dataset contains two examples. As observed, when ConfExplainer produces accurate explanations, the confidence score is higher. Conversely, for cases where the explanation is not entirely correct, the confidence score is lower. This demonstrates the ability of the confidence module to provide a more reliable assessment of the accuracy of generated explanations.

% \begin{figure*}[]
%     \centering
%     \includegraphics[width=1.0\textwidth]{figures/case_study/conf_all.png}
%     \caption{Case study}
%     \label{fig:enter-label}
% \end{figure*}